%% file: main.tex
\documentclass[10pt,twocolumn,letterpaper]{article}

\usepackage{cvpr}              

\usepackage{amsmath,amsfonts}
\usepackage{amssymb} 

\usepackage{array}

\usepackage{textcomp}
\usepackage{stfloats}
\usepackage{verbatim}
\usepackage{graphicx}

\usepackage{multirow}

\usepackage{xcolor}
\usepackage{colortbl}

\usepackage{makecell}
\usepackage{overpic}
\usepackage{rotating,soul}
\usepackage{bbding}
\usepackage{ragged2e}
\usepackage{booktabs}
\usepackage[normalem]{ulem}
\usepackage{float}
\usepackage{stfloats}
\usepackage{bbold}
\usepackage{bbm}
\usepackage{silence}
\usepackage{pifont}
\usepackage[ruled,vlined]{algorithm2e}
\usepackage{algorithmic,algorithm2e,float}
\SetKwInput{KwInput}{Input}
\SetKwInput{KwOutput}{Output} 

\def\OURS{GET}
\definecolor{mygray}{gray}{.92}
\newcommand{\myPara}[1]{\vspace{0.04in}\noindent\textbf{#1}\quad}
\def\ie{\emph{i.e.}}

\def\etal{{\em et al.~}}
\newcommand{\cmark}{\ding{51}}%
\newcommand{\xmark}{\ding{55}}%

\usepackage[accsupp]{axessibility}


\definecolor{cvprblue}{rgb}{0.21,0.49,0.74}
\usepackage[pagebackref,breaklinks,colorlinks,citecolor=cvprblue]{hyperref}

\def\paperID{15059} 
\def\confName{CVPR}
\def\confYear{2025}

\title{GET: Unlocking the Multi-modal Potential of CLIP for \\ Generalized Category Discovery}

\author{Enguang Wang$^1$, Zhimao Peng$^1$, Zhengyuan Xie$^1$, Fei Yang$^{2,1}$\thanks{Corresponding author.}, Xialei Liu$^{2,1}$, Ming-Ming Cheng$^{2,1}$\\
$^1$VCIP, CS, Nankai University \qquad $^2$NKIARI, Shenzhen Futian\\ 
{\tt\small \{enguangwang,zhimao796,xiezhengyuan\}@mail.nankai.edu.cn} \\
{\tt\small \{feiyang,xialei,cmm\}@nankai.edu.cn}
}

\begin{document}
\maketitle
\input{sec/0_abstract}

\input{sec/1_intro}
\input{sec/2_related_works}

\input{sec/3_preliminary}
\input{sec/4_method}
\input{sec/5_experiments}
\input{sec/6_conclusion}

\clearpage
\input{sec/7_ack}
{
    \small
    \bibliographystyle{ieeenat_fullname}
    \bibliography{main}
}
\input{sec/X_suppl}

\end{document}

%% file: sec/0_abstract.tex
\begin{abstract}
Given unlabelled datasets containing both old and new categories, generalized category discovery (GCD) aims to accurately discover new classes while correctly classifying old classes.  
Current GCD methods only use a single visual modality of information, resulting in a poor classification of visually similar classes. 
As a different modality, text information can provide complementary discriminative information, which motivates us to introduce it into the GCD task.
However, the lack of class names for unlabelled data makes it impractical to utilize text information.
To tackle this challenging problem, in this paper, we propose a Text Embedding Synthesizer (TES) to generate pseudo text embeddings for unlabelled samples. Specifically, our TES leverages the property 
that CLIP can generate aligned vision-language features, 
converting visual embeddings into tokens of the CLIP’s text 
encoder to generate pseudo text embeddings. Besides, we employ a dual-branch framework, through the joint learning and instance consistency of different modality branches, visual
and semantic information  mutually enhance each other,
promoting the interaction and fusion
of visual and text knowledge.
Our method unlocks the multi-modal potentials of CLIP and outperforms the baseline methods by a
large margin on all GCD benchmarks, achieving new state-of-the-art. Our code is available at: \url{https://github.com/enguangW/GET}. 
\end{abstract}

%% file: sec/1_intro.tex
\section{Introduction}
\label{sec:intro}

Deep neural networks trained on large amounts of labeled data have shown powerful visual recognition capabilities~\cite{krizhevsky2017imagenet}. Although this is heartening, the close-set assumption
severely hinders the deployment of the model in practical application scenarios.
Recently, novel class discovery (NCD)~\cite{han2019learning} has been proposed to categorize unknown classes of unlabelled data, leveraging knowledge learned from labeled data.
As a realistic extension to NCD,
generalized category discovery (GCD) ~\cite{vaze2022generalized} assumes that the
unlabelled data come from both known and unknown
classes, rather than just unknown classes as in NCD. The model needs to accurately discover unknown classes while correctly classifying known classes of the unlabelled data,
breaking the close-set limitation, making GCD a challenging and meaningful task.

\begin{figure}[t]
    \centering
    \includegraphics[width=0.98\columnwidth]{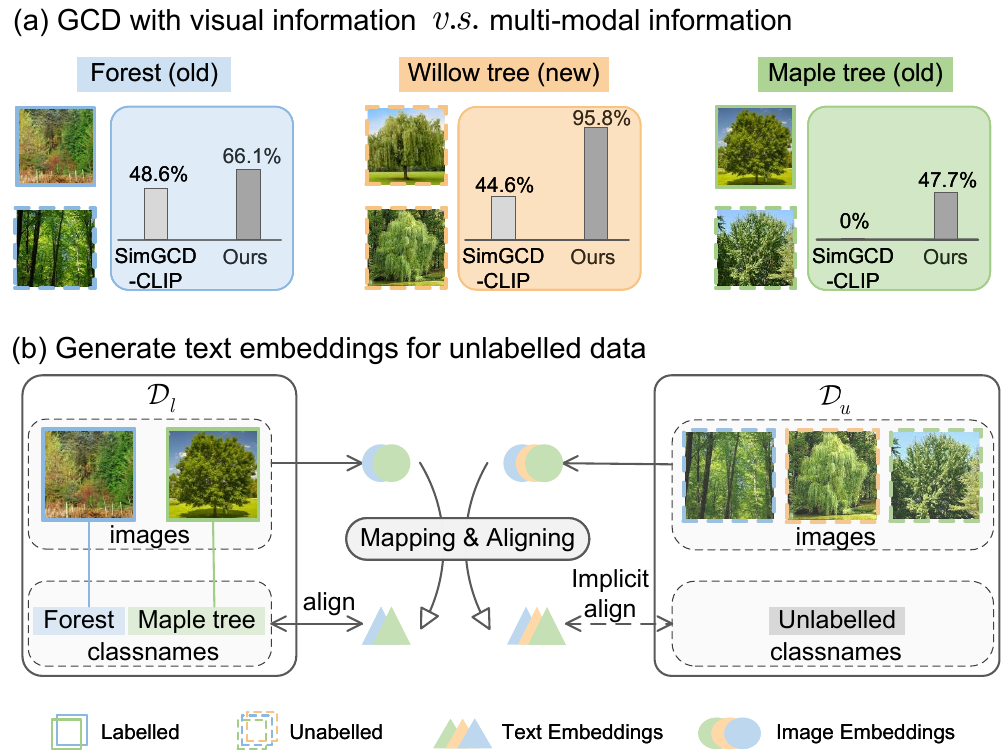}
    \caption{The motivation of our method. (a) Current GCD methods~\cite{wen2023parametric} rely on single visual features, resulting in poor classification of visually similar classes; 
    Our approach introduces text information, improving the discriminative capabilities of the model. (b) Our proposed method maps image embeddings to text embeddings while simultaneously achieving modal alignment.
    }
    \label{fig:teaser}
    \vspace{-6mm}
\end{figure}

Previous GCD methods~\cite{vaze2022generalized,wen2023parametric,zhang2023promptcal,pu2023dynamic,zhao2023learning}
utilize a DINO~\cite{caron2021emerging} pre-trained ViT as the backbone network to expect good initial discrimination ability of the model, thereby facilitating fine-tuning on the training data. 
Although promising results have been achieved, these representations derived from a single visual backbone often struggle with visually similar categories, such as the classes in all fine-grained datasets and some super-class subsets of generic datasets. 
As shown in \cref{fig:teaser} (a), replacing the backbone of the parametric baseline~\cite{wen2023parametric} with the powerful CLIP~\cite{radford2021learning} visual encoder still struggles to generalize certain visual concepts, leading to sub-optimal results.
Inspired by the idea that the textual modality can provide complementary discriminative information, we decided to introduce text information into the GCD task to compensate for the insufficient discriminative of visual concepts.
However, the lack of class names for unlabelled data in GCD makes it impractical to use the text encoder, thus locking the multi-modal potential of CLIP for GCD.

In order to tackle this challenging problem, in this paper, we propose a generative-based method to \textbf{GE}nerate pseudo \textbf{T}ext embeddings for unlabelled data, dubbed \textbf{\OURS}.
In particular, we first introduce a 
\textit{{T}ext {E}mbedding {S}ynthesizer} ({TES}) module based on the vision-language alignment property of CLIP, producing reliable and modality-aligned pseudo text features. 
As shown in \cref{fig:teaser} (b),
TES learns a mapping that transforms image embeddings into text embeddings. 
Specifically, TES converts visual embeddings into tokens for the text encoder, eliminating the need for textual input. To mitigate the gap between generated pseudo-text embeddings and real text embeddings, TES distills knowledge from real text embeddings corresponding to labeled data. Additionally, TES aligns text and image for the same instance, enforcing consistency between language and vision while preventing overfitting to known classes.
This training approach renders TES equivalent to a special finetuned text encoder with only visual input. From another perspective, our TES can be considered as performing an image captioning task~\cite{merullo2022linearly}.

To leverage such multi-modal features in the GCD task, we propose 
a dual-branch multi-modal joint training strategy with a cross-modal instance consistency objective. One branch focuses on visual information, while the other branch supplements it with text information. Through joint learning on the GCD task, visual and semantic information aspects mutually enhance each other. 
Furthermore, our cross-modal instance consistency objective 
enforces  the  instance
have the same relationship in both visual and text modalities with anchors constructed by labeled instances, promoting the interaction and alignment of visual and text embedding space. 
With the supplementation of text embeddings generated by TES and an appropriate dual-branch training strategy, the multi-modal features correct the classification hyperplane, enhancing discriminative ability while 
reducing bias issues.

To summarize, our contributions are as follows:
\begin{itemize}
\item To tackle the problem that the text encoder can not be used on the unlabelled data,
we propose a TES module converting visual embeddings into tokens of the CLIP’s text encoder to generate pseudo text embeddings.
\item Through the proposed cross-modal instance consistency objective in our dual-branch framework, information of different modalities mutually enhances each other, producing more discriminative classification prototypes.
\item Our method achieves state-of-the-art results on
multiple benchmarks, providing GCD  a multi-modal paradigm.
\end{itemize}

%% file: sec/2_related_works.tex
\section{Related Works}
\label{sec:rel_works}

\myPara{Novel Class Discovery (NCD).}
NCD can be traced back to KCL~\cite{hsu2018learning}, where pairwise similarity generated by a similarity prediction network guides clustering reconstruction, offering a constructive approach for transfer learning across tasks and domains. 
Early methods are based on two objectives: pretraining on labeled data and clustering on unlabelled data.
RS~\cite{Han2020automatically} performs a self-supervised pretraining on both labeled and unlabelled data, alleviating the model's bias towards known classes. Simultaneously,  RS  proposes knowledge transfer through rank statistics, which has been widely adopted in subsequent research. \cite{zhao21novel} proposes a two-branch learning framework with dual ranking statistics, exchanging information through mutual knowledge distillation, which is similar to our approach to some extent. Differently, our two branches focus on semantic and visual information rather than local and global characteristics in~\cite{zhao21novel}. In order to simplify NCD approaches, UNO~\cite{fini2021unified} recommends optimizing the task with a unified cross-entropy loss using the multi-view SwAV~\cite{caron2020unsupervised} exchange prediction strategy, which sets a new paradigm. 

\myPara{Generalized Category Discovery (GCD).}
Recently, GCD~\cite{vaze2022generalized} extends  NCD to a more realistic scenario, where unlabelled data comes from both known and unknown classes. GCD~\cite{vaze2022generalized} employs a pre-trained vision transformer~\cite{dosovitskiy2021an} to provide initial visual representations, fine-tuning the backbone through supervised and self-supervised contrastive learning on the labeled and the entire data. Once the model learns discriminative representations, semi-supervised k-means are used for classification by constraining the correct clustering of labeled samples. As an emerging and realistic topic, GCD is gradually gaining attention.  PromptCAL~\cite{zhang2023promptcal} propose a two-stage
framework to tackle the class collision issue caused by false negatives while enhancing the adaptability of the model on downstream datasets. 
SimGCD~\cite{wen2023parametric} introduces a parametric classification approach, addressing the computational overhead of GCD clustering while achieving remarkable improvements. Specifically, SimGCD adds a classifier on top of GCD and jointly learns self-distillation and supervised training strategies.
$\mu$GCD~\cite{mugcd} examines the taxonomy bias in previous methods by introducing the Clevr4 dataset and employs the mean-teacher technique and a more efficient training strategy thus achieving remarkable performance improvements.  
CLIP-GCD~\cite{ouldnoughi2023clip} mines text descriptions from a large text corpus to use the text encoder and simply concatenates visual and text features for classification. In contrast, our method focuses on the dataset itself, without introducing additional corpus. 
Most recently, TextGCD~\cite{zheng2024textual} collects many text tags from multiple benchmarks and leverages LLMs to enhance these tags, constructing a visual lexicon. It then generates textual descriptions for each sample based on the similarity between the visual lexicon and visual feature.
Different from these methods, our GET employs CLIP to introduce multi-modal information into the task without relying on any additional databases or large models. 

\myPara{Vision-Language Pre-training.}
Vision-Language pre-training~\cite{du2022survey,chen2023vlp, li2022blip, MIR-2022-06-193, chen2024far, gao2024mini} aims to train a large-scale model on extensive image-text data, which, through fine-tuning, can achieve strong performance on a range of downstream visual-language tasks. 
Some studies~\cite{li2019visualbert,chen2020uniter,lu2019vilbert,tan2019lxmert,li2021align} achieve improved performance in various image-language tasks by modeling image-text interactions through a fusion approach. 
However, the need to encode all image-text pairs in the fusion approach makes the inference speed in image-text retrieval tasks slow. Consequently, some studies~\cite{radford2021learning,jia2021scaling} propose 
a separate encoding of images and texts, and project image and text embeddings into a joint embedding space
through contrastive learning.
CLIP~\cite{radford2021learning} uses contrastive training on large-scale image-text pairs, minimizing the distance between corresponding images and texts while simultaneously maximizing the distance between non-corresponding pairs.
The strong generalization capabilities and multi-modal properties of CLIP prompt us to introduce it to the GCD Task.

%% file: sec/3_preliminary.tex
\section{Preliminaries}
\label{sec:gem_GCD}

\subsection{Problem formulation}
In the context of GCD, the training data~$\mathcal{D}$ is divided into a labeled dataset~$\mathcal{D}_l=\left\{(\boldsymbol{x}_i^l, {y}_i^l)\right\}_{i=1}^N\in \mathcal{X}\times \mathcal{Y}_l$ and an unlabelled dataset
$\mathcal{D}_u=\left\{(\boldsymbol{x}_i^u, {y}_i^u)\right\}_{i=1}^M\in \mathcal{X}\times \mathcal{Y}_u$, where $\mathcal{Y}_l$ and $\mathcal{Y}_u$ represent the label space while
$\mathcal{Y}_l \subset \mathcal{Y}_u$, and $\mathcal{D} = \mathcal{D}_l \cup \mathcal{D}_u $. 
$|\mathcal{Y}_l|$ and $|\mathcal{Y}_u|$ represent the number of categories for labeled samples and unlabelled samples, respectively. 
Following the setting in~\cite{vaze2022generalized,wen2023parametric}, we assume the class number of new classes $|\mathcal{Y}_u\backslash\mathcal{Y}_l|$ is known, or it can be estimated through some off-the-shelf methods~\cite{vaze2022generalized,han2019learning}.
The goal of GCD is to correctly cluster unlabelled samples with the help of labeled samples. 

\subsection{Parametric GCD method (SimGCD)}
\label{ssec:simgcd}
In this paper, we tackle the GCD problem in a parametric way which is proposed by SimGCD \cite{wen2023parametric}. It trains a unified prototypical classification head for all new/old classes to perform GCD through a DINO-like form of self-distillation. Specifically, it includes two types of loss functions: representation learning and parametric classification. For representation learning, it performs supervised representation learning \cite{khosla2020supervised} $\mathcal{L}_{scon}$ on all labeled data and self-supervised contrastive learning $\mathcal{L}_{con}$ on all training data, the loss functions are as follows:
\begin{equation}\label{eq:s_con}
\mathcal{L}_{scon}=-\frac{1}{|B_l|} \sum_{i \in B_l}\frac{1}{|\mathcal{N}_i|} \sum_{q \in \mathcal{N}_i} \log \frac{\exp \left(\frac{{\boldsymbol{h}_i}^\top {\boldsymbol{h}_q}^{\prime}}{{\tau_{sc}}}\right)}{\sum_{n\in B_l}^{n \ne i} \exp \left(\frac{{\boldsymbol{h}_i}^\top {\boldsymbol{h}_n}^{\prime}}{\tau_{sc}}\right)} \,,
\end{equation}
\begin{equation}\label{eq:con}
\mathcal{L}_{con}=-\frac{1}{|B|} \sum_{i \in B}\log \frac{\exp \left(\frac{{\boldsymbol{h}_i}^\top {\boldsymbol{h}_i}^{\prime}}{{\tau_c}}\right)}{\sum_{n\in B}^{n \ne i} \exp \left(\frac{{\boldsymbol{h}_i}^\top {\boldsymbol{h}_n}^{\prime}}{\tau_c}\right)} \,,
\end{equation}
 where  $\mathcal{N}_i$ denotes the indices of other images with the same semantic label as $\boldsymbol{x}_i$ in a batch,
$B_l$ corresponds to the labeled subset of the mini-batch $B$, $\tau_{sc}$ and $\tau_c$ are temperature values. For visual embeddings  $\boldsymbol{z}_i$ and $\boldsymbol{z}_i^\prime$ of two views $\boldsymbol{x}_i$ and $\boldsymbol{x}_i^\prime$ generated by the image encoder, an MLP layer  $g(\cdot)$ is used to map  $\boldsymbol{z}_i$ and $\boldsymbol{z}_i^\prime$ to high-dimensional embeddings $\boldsymbol{h}_i=g(\boldsymbol{z}_i)$ and $\boldsymbol{h}_i^\prime=g(\boldsymbol{z}_i^\prime)$. For parametric classification, all labeled data are trained by a cross-entropy loss $\mathcal{L}_\text{cls}^s$ and all training data are trained by a self-distillation loss $\mathcal{L}_\text{cls}^u$:
\begin{equation}\label{eq:cls}
\mathcal{L}_\text{cls}^s = \frac{1}{|B_l|} \sum_{i \in B^l} \mathcal{H}\left(y_i,\sigma(\boldsymbol{p}_i, \tau_s)\right) \,,
\end{equation}
\begin{equation}\label{eq:dis}
\mathcal{L}_\text{cls}^u = \frac{1}{|B|} \sum_{i \in B} \mathcal{H}\left(\sigma({\boldsymbol{p}_i}^\prime, \tau_t),\sigma(\boldsymbol{p}_i, \tau_s)\right) \,,
\end{equation}
where $\sigma(\cdot)$ is the softmax function, $\boldsymbol{p}_i$ and ${\boldsymbol{p}_i}^\prime$ are the outputs of two views $x_i$ and $x_i^\prime$ on the prototypical classifier, respectively. $\tau_s$ is a temperature parameter and $\tau_t$ is a sharper version. $\mathcal{H}(\cdot)$ denotes the cross-entropy function, $y_i$ is the corresponding ground truth of $x_i$, and  $\sigma({\boldsymbol{p}_i}^\prime, \tau_t)$ is the soft pseudo-label of $x_i$.

In addition, SimGCD also introduces a mean-entropy maximization regularization term $H(\overline{\boldsymbol{p}})$  to prevent trivial solutions, where $H(\cdot)$ is the entropy of predictions \cite{shannon1948mathematical}, $\overline{\boldsymbol{p}}=\frac{1}{2|B|} \sum_{i \in B}\left(\sigma({\boldsymbol{p}_i}^\prime, \tau_s) +\sigma(\boldsymbol{p}_i, \tau_s)\right)$ is the mean softmax probability of a batch. By using the above loss functions and regularization term to train the model, SimGCD has achieved significant improvements, however, it struggles with performance on visually similar categories due to the use of single visual modality information.

%% file: sec/4_method.tex
\section{Our Method}
In this paper, we propose \OURS, which addresses the GCD task in a multi-modal paradigm. 
As shown in \cref{fig:method}, our~\OURS~contains two stages. In the first stage, we learn a text embedding synthesizer (TES, in \cref{ssec:tes}) to
generate pseudo text embeddings for each sample. In the second stage, a dual-branch multi-modal joint training strategy with cross-modal instance consistency (in \cref{ssec:DCGC}) is introduced to fully leverage multi-modal features.
\subsection{Text embedding synthesizer}
\label{ssec:tes}
The absence of natural language class names for unlabelled data makes it challenging to introduce text information into the GCD task. 
In this paper, we attempt to generate pseudo text embeddings aligned with visual embeddings for each image from a feature-based perspective.

\begin{figure*}[t]
    \centering
    \includegraphics[width=\textwidth]{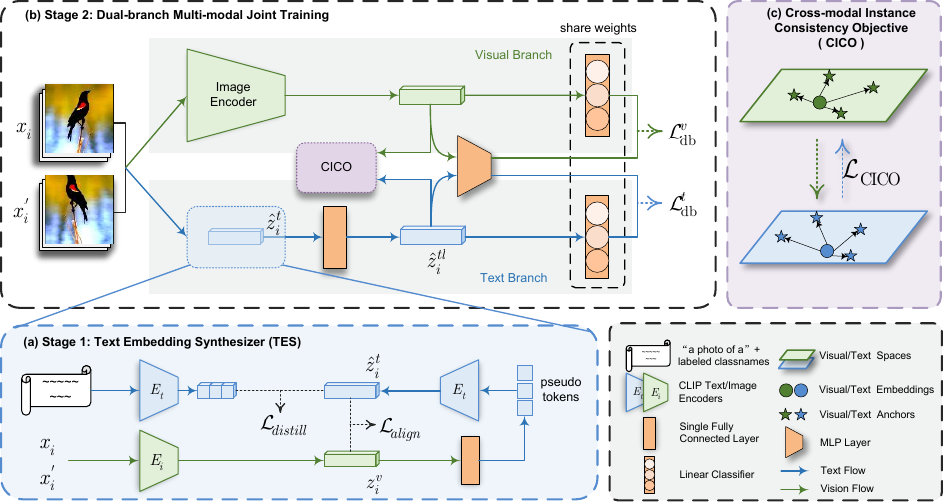} 
    \caption{Overview of our~\OURS          ~framework. (a) In the first stage, we introduce a text embedding synthesizer that generates pseudo text embeddings for unlabelled data. TES learns a linear mapping that transforms image features into input tokens for the text encoder. The resulting pseudo text embeddings are then used for joint training in the second stage. (b) We proposed a dual-branch multi-modal joint training framework with a cross-modal instance consistency objective in the second stage. Two branches utilize the same parameterized training strategy~\cite{wen2023parametric} while focusing on text and visual information, respectively. (c) Our cross-modal instance consistency objective
    makes visual and text information
  exchange and benefit from each other.
    }
    \label{fig:method}
    \vspace{-6mm}
\end{figure*}

Inspired by BARON~\cite{wu2023baron}, which treats embeddings within bounding boxes as embeddings of words in a sentence to solve the open-vocabulary detection task, we propose a text embedding synthesizer (TES). Specifically, our TES leverages the property that CLIP can generate aligned vision-language features, converting visual embeddings into tokens of the CLIP's text encoder to generate pseudo text embeddings for each sample. The architecture of TES is shown in \cref{fig:method} (a). 
For each image $\boldsymbol{x}_i$ in a mini-batch, we use CLIP's image encoder to obtain its visual embedding $\boldsymbol{z}_i^v$. A single fully connected layer $l$ is used to map the visual embedding to pseudo tokens that serve as input to the CLIP's text encoder, thus generating corresponding pseudo text embeddings $\boldsymbol{\hat{z}}_i^t$.

The objective of TES contains an align loss on all samples and a distill loss on labeled samples.
To align the generated pseudo text embeddings $\boldsymbol{\hat{z}}_i^t$  with their corresponding visual features $\boldsymbol{z}_i^v$, our align loss leverages the modality alignment property of CLIP's encoders, pulling correct visual-text embedding pairs closer while pushing away the incorrect ones.
The align loss  consists of symmetric components $\mathcal{L}_{align}^v$ and $\mathcal{L}_{align}^t$, 
calculated as:
\begin{equation}
\label{eq:align1}
\mathcal{L}_{align}^v = -\frac{1}{|B|}\sum_{i \in B}\log \frac{\exp \left({\boldsymbol{z}_i^v}^\top {\boldsymbol{\hat{z}}_i^t}/ \tau_a\right)}{\sum_{j \in B}   \exp \left({\boldsymbol{z}_i^v}^\top {\boldsymbol{\hat{z}}_j^t}/ \tau_a\right)} \,,
\end{equation}
\begin{equation}
\label{eq:align2}
\mathcal{L}_{align}^t = -\frac{1}{|B|}\sum_{i \in B}\log \frac{\exp \left( {\boldsymbol{\hat{z}}_i^t}^\top{\boldsymbol{z}_i^v}/ \tau_a\right)}{\sum_{j \in B}   \exp \left({\boldsymbol{\hat{z}}_i^t}^\top{\boldsymbol{z}_j^v}/ \tau_a\right)} \,,
\end{equation}
where 
$\boldsymbol{\hat{z}}_i^t$ and $\boldsymbol{z}_i^v$ are $\ell_2$-normalised,
and $\tau_a$ is a temperature parameter. Thus, the align loss is
$\mathcal{L}_{align}= \mathcal{L}_{align}^v + \mathcal{L}_{align}^t$.

To ensure that our generated pseudo text features reside in the same embedding space as real text features and maintain consistency, we introduce a distill loss $\mathcal{L}_{distill}$:
\begin{flalign}
\label{eq:distill}
\begin{split}
\mathcal{L}_{distill}= & -\frac{1}{|B_l|}\sum_{i \in B_l}\log\frac{\exp \left({{\boldsymbol{\hat{z}}_i^t}^\top{T(n_i)}} \right)}{\sum_{j=0}^{|\mathcal{Y}_l|}\mathbbm{1}_{[j\ne n_i]}   \exp \left({\boldsymbol{\hat{z}}_i^t}^\top{T(j)}\right)} \\
& + \frac{1}{|B_l|}\sum_{i \in B_l} \left( \boldsymbol{\hat{z}}_i^t - T(n_i)\right)^2  \,,
\end{split}
\end{flalign}
where $T \in |\mathcal{Y}_l| \times dim$ are the real text embeddings of $|\mathcal{Y}_l|$  semantic labels, $n_i \in \mathcal{Y}_l$ indexes the corresponding class name of $\boldsymbol{x}_i$ among all known class names,
$T(j)$ denotes the $j$-th real text embeddings of all known class names and $\mathbbm{1}_{[\cdot]}$ is the indicator function. Each vector in $T$ is produced by the text encoder using the prompt ``a photo of a \{CLS\}'' where \{CLS\} denotes the corresponding class name.


The overall objective of our text embedding synthesizer is $
\mathcal{L}_{TES}= \mathcal{L}_{align} + \mathcal{L}_{distill} $.
The distill loss is used to guide pseudo text embeddings of the network's output towards the real semantic corresponding space and adapt the model to the distribution of the dataset, while the align loss prevents overfitting to known classes and enforces the consistency between two modalities. 
Moreover, we introduce a multi-view strategy for TES. Specifically, we calculate both the align loss $\mathcal{L}_{align}$ and distill loss $\mathcal{L}_{distill}$ for two different views $\boldsymbol{x}_i$ and ${\boldsymbol{x}_i}^\prime$ of the same image in a mini-batch. This further implicitly enhances the instance discriminative nature~\cite{wu2018unsupervised} of our TES training, allowing different views of the same labeled image to generate identical pseudo text embeddings. The generated pseudo text embeddings $\boldsymbol{\hat{z}}_i^t$ are then used for joint training in the second stage.

\subsection{Dual-branch multi-modal joint training}
\label{ssec:DCGC}
Intuitively, the introduction of multi-modal information can have a positive impact on the GCD task.
Textual information can serve as an effective complement to visual information, enhancing the model's discriminative capabilities.
However, how to effectively utilize visual and text information in the GCD task and make the most of their respective roles remains challenging. 
In this paper, we propose a dual-branch architecture as illustrated in \cref{fig:method} (b),
which focuses on semantic and visual information, respectively.
We employ the same parametric training strategy (in \cref{ssec:simgcd}) for each branch to promote that the model has aligned and complementary discriminative capabilities for visual and text features of the same class. Furthermore, we introduce a cross-modal instance consistency loss, which constrains the instance relationships of samples in both visual and text spaces, enabling the two branches to learn from each other. We use the indicator $v$ to represent the visual concept while $t$ for the text concept.

\myPara{Visual-branch.} The objective of the visual branch contains a representation learning part and a parametric classification part. Given a visual embedding $\boldsymbol{z}_i^v$  of image $\boldsymbol{x}_i$ generated by the image encoder, an MLP layer  $g(\cdot)$ is used to map  $\boldsymbol{z}_i^v$ to a high-dimensional embedding $\boldsymbol{h}_i^v=g(\boldsymbol{z}_i^v)$. Meanwhile, we employ a prototypical classifier $\eta(\cdot)$ to generate classification probability distribution $\boldsymbol{p}_i^v = \eta(\boldsymbol{z}_i^v)$.
Simply replace all the high-dimensional embedding $\boldsymbol{h}$ (the subscript  is omitted for brevity) in \cref{eq:s_con} and \cref{eq:con} with its corresponding visual branch version $\boldsymbol{h}^v$ can obtain the the supervised contrastive loss $\mathcal{L}_{scon}^v$
 and the self-supervised contrastive loss $\mathcal{L}_{ucon}^v$. The overall representation learning loss is balanced with $\lambda$, written as:
\begin{equation}
\label{eq:total_loss}
\mathcal{L}_\text{rep}^v = (1 - \lambda) \mathcal{L}_{ucon}^{v} +
\lambda  \mathcal{L}_{scon}^{v}
\,.
\end{equation}

For the parametric classification part, just replace $\boldsymbol{p}_i$ and ${\boldsymbol{p}_i}^\prime$ of  \cref{eq:cls} and \cref{eq:dis} with $\boldsymbol{p}_i^v$ and ${\boldsymbol{p}_i^v}^\prime$ can obtain the cross-entropy loss $\mathcal{L}_\text{cls-v}^s$ and the self-distillation loss $\mathcal{L}_\text{cls-v}^u$. Thus, the classification loss is $
\mathcal{L}_\text{cls}^v = (1 - \lambda)\mathcal{L}_\text{cls-v}^u +
\lambda \mathcal{L}_\text{cls-v}^s
$.

The overall objective of the visual branch is as follows:
\begin{equation}
\label{eq:b_v}
\mathcal{L}_\text{db}^v =  \mathcal{L}_\text{rep}^v + \mathcal{L}_\text{cls}^v\,.
\end{equation}
\myPara{Text-branch.}
Our text branch simply adopts the same training strategy as the visual branch. That is, in particular, given a text embedding $\boldsymbol{\hat{z}}_i^t$ generated by TES, we first  input it into 
 a fully connected layer to gain  a learnable text embedding $\boldsymbol{\hat{z}}_i^{tl}$ while change its dimension.
Simply replace  $\boldsymbol{h}_i^v$ in the representation learning objective $\mathcal{L}_\text{rep}^v$ with $\boldsymbol{h}_i^t=g(\boldsymbol{\hat{z}}_i^{tl})$
and replace ${\boldsymbol{p}_i^v}$ in
classification parts
$\mathcal{L}_\text{cls}^v$ with  
 $\boldsymbol{p}_i^t = \eta(\boldsymbol{\hat{z}}_i^{tl})$
can yield the corresponding text objectives  $\mathcal{L}_\text{rep}^t$ and $\mathcal{L}_\text{cls}^t$. In other words, changing the visual conception indicator $v$ into text conception indicator $tl$ can get the corresponding objectives for the text branch.
Thus, the overall objective of our text branch can be formally written as $\mathcal{L}_\text{db}^t =  \mathcal{L}_\text{rep}^t + \mathcal{L}_\text{cls}^t$

To mitigate the bias between old and new classes, we extended the mean-entropy regularization~\cite{assran2022masked} to a multi-modal mean entropy regularization
$H_{mm}={H}(\overline{p}_{mm},\overline{p}_{mm}) $, here $\overline{p}_{mm}$ can calculate by $\overline{p}_{mm} = \frac{1}{2|B|}\sum_{i \in B}\left( \sigma({\boldsymbol{p}_i^v}, \tau_s) +\sigma(\boldsymbol{p}_i^t, \tau_s) \right)$.
In this way, the prediction probabilities in different modalities for each prototype are constrained to be the same, preventing trivial solutions.

\myPara{Cross-modal instance 
consistency objective.}
In order to enable the two branches to learn from each other while encouraging agreement between two different modes, we propose a cross-modal instance consistency objective (CICO), shown in \cref{fig:method} (c). Our CICO has the same form of mutual knowledge distillation as ~\cite{zhao21novel}, but we distill the instance consistency between the two branches. For each mini-batch $B$, we choose its labeled subset $B_l$ containing $K$ categories as anchor samples, calculate the visual and text prototypes for $K$ categories as visual anchors $\mathcal{P}_v$ and text anchors $\mathcal{P}_t$, respectively. 
We define the instance relationships in visual and text branches as:
\begin{align}
\label{eq:anchor}
\begin{split}
s_i^v = & \sigma({\boldsymbol{z}_i^v}^\top\mathcal{P}_v)\,, \\
s_i^t = & \sigma({\boldsymbol{\hat{z}}_i^{tl}}^\top\mathcal{P}_t) \,.
\end{split}
\end{align}
Thus our CICO can formally written as:
\begin{equation}
\label{eq:cigc}
\mathcal{L}_\text{CICO} =  \frac{1}{2|B|}\sum_{i \in B}\left(D_{KL}(s_i^t||s_i^v) + D_{KL}(s_i^v||s_i^t)
\right)\,,
\end{equation}
where $D_{KL}$ is the  Kullback-Leibler divergence.
Mutual knowledge distillation on instance relationships for two modalities makes visual and text flows exchange and benefit from each other, thus the two branches can serve as complementary discriminative aids to each other.

The overall optimization objective of our method is:
\begin{equation}
\label{eq:db}
\mathcal{L}_\text{Dual} =  \mathcal{L}_\text{db}^v + \mathcal{L}_\text{db}^t - \epsilon H_{mm} + \lambda_c\mathcal{L}_\text{CICO} \,.
\end{equation}
Since information
from different modalities is exchanged and learned through CICO and injected
into the visual backbone, we utilize the last-epoch visual branch for
inference. 

%% file: sec/5_experiments.tex
\section{Experiments}
\label{sec:exp}
\subsection{Experimental setup}
\label{ssec:es}

\begin{table*}[!tp]
\centering

\setlength{\tabcolsep}{3pt}
\resizebox{\textwidth}{!}{%
\begin{tabular}{lcccccccccccccccccc}
\toprule
&   \multicolumn{3}{c}{CUB} & \multicolumn{3}{c}{Stanford Cars} & \multicolumn{3}{c}{FGVC-Aircraft} &    \multicolumn{3}{c}{CIFAR10} & \multicolumn{3}{c}{CIFAR100} & \multicolumn{3}{c}{ImageNet-100}
\\
\cmidrule(rl){2-4}\cmidrule(rl){5-7}\cmidrule(rl){8-10}\cmidrule(rl){11-13}\cmidrule(rl){14-16}\cmidrule(rl){17-19}
Method        & All  & Old  & New  & All  & Old  & New  & All  & Old  & New & All  & Old  & New  & All  & Old  & New  & All  & Old  & New  \\
\midrule
$k$-means~\cite{macqueen1967some}  & 34.3 & 38.9 & 32.1 & 12.8 & 10.6 & 13.8 & 16.0 & 14.4 & 16.8 & 83.6 & 85.7 & 82.5 & 52.0 & 52.2 & 50.8 & 72.7 & 75.5 & {71.3} \\
RS+~\cite{Han2020automatically}          & 33.3 & 51.6 & 24.2 & 28.3 & 61.8 & 12.1 & 26.9 & 36.4 & 22.2 & 46.8 & 19.2 & 60.5 & 58.2 & {77.6} & 19.3 & 37.1 & 61.6 & 24.8 \\
UNO+~\cite{fini2021unified}               & 35.1 & 49.0 & 28.1 & 35.5 & 70.5 & 18.6 & 40.3 & 56.4 & 32.2 & 68.6 & \textbf{98.3} & 53.8 & 69.5 & 80.6 & 47.2 & 70.3 & {95.0} & 57.9 \\
ORCA~\cite{cao2022openworld}                     & 35.3 & 45.6 & 30.2 & 23.5 & 50.1 & 10.7 & 22.0 & 31.8 & 17.1 & 81.8 & 86.2 & 79.6 & 69.0 & 77.4 & 52.0 & 73.5 & {92.6} & 63.9  \\
\midrule

GCD~\cite{vaze2022generalized}       & {51.3} & {56.6} & {48.7} & {39.0} & 57.6 & {29.9} & {45.0} & 41.1 & {46.9}& {91.5} & {97.9} & {88.2} & {73.0} & 76.2 & {66.5} & {74.1} & 89.8 & 66.3 \\
GPC~\cite{Zhao_2023_ICCV}       & {55.4} & {58.2} & {53.1} & {42.8} & 59.2 & {32.8} & {46.3} & 42.5 & {47.9}& {92.2} & {98.2} & { 89.1} & {77.9} & 85.0 & {63.0} & {76.9} & 94.3 & { 71.0} \\
DCCL~\cite{dccl} & {63.5} & {60.8} & {64.9} & {43.1} & 55.7 & {36.2} & {-} &- & {-} & {96.3} & {96.5} & {96.9} & { 75.3} & 76.8 & {70.2} & {80.5} &90.5 & {76.2}\\
PromptCAL~\cite{zhang2023promptcal} & {62.9} & {64.4} & {62.1} & {50.2} & 70.1 & {40.6} & {52.2} &52.2 & {52.3}& \textbf{97.9} & {96.6} & {98.5} & {81.2} & 84.2 & {75.3} & {83.1} &92.7 & {78.3} \\
SimGCD~\cite{wen2023parametric}                    & 
60.3 & 65.6 & 57.7 & 53.8 & 71.9 & 45.0 & 54.2 & 59.1 & 51.8 & 97.1 & 95.1 & 98.1 & 80.1 & 81.2 & {77.8} & 83.0 & 93.1 & 77.9 \\
$\mu$GCD~\cite{mugcd}  &  65.7 & 68.0&  64.6&  56.5&  68.1 & {50.9}&  53.8 & 55.4 & 53.0  & {-} &- & {-} & {-} &- & {-} & {-} &- & {-}\\
LegoGCD~\cite{legogcd}& 63.8 &71.9 &59.8 &57.3& 75.7 &48.4 &55.0& 61.5 &51.7 &97.1& 94.3& {98.5}& 81.8 &81.4 &\textbf{82.5}& {86.3} & {94.5} &{ 82.1}

\\
\midrule

GCD-CLIP       & {57.6} & {65.2} & {53.8} & {65.1} & 75.9 & {59.8} & {45.3} & 44.4 & {45.8} & {94.0} & {97.3} & {92.3} & {74.8} & 79.8 & {64.6} & {75.8} & 87.3 & 70.0 \\

SimGCD-CLIP                    & 71.7 & 76.5 & 69.4 & 70.0 & 83.4 & 63.5 & 54.3 & 58.4 & 52.2& 97.0 & {94.2} & {98.4} & {81.1} & {85.0} & {73.3} & {90.8} & {95.5} & {88.5}  \\
\midrule
\rowcolor{mygray}
\textbf{\OURS} (Ours)                    & \textbf{77.0} & \textbf{78.1} & \textbf{76.4} & \textbf{78.5} & \textbf{86.8} & \textbf{74.5} & \textbf{58.9} & \textbf{59.6} & \textbf{58.5}  & {97.2} & {94.6} & \textbf{98.5} & \textbf{82.1} & \textbf{85.5} & {75.5} & \textbf{91.7} & \textbf{95.7} & \textbf{89.7}\\
\bottomrule
\end{tabular}}
\caption{Results (\%) on fine-grained and generic datasets. The best results are highlighted in \textbf{bold}.} 
\label{subtab:ssb}
\vspace{-5mm}
\end{table*}

\myPara{Datasets.} We evaluate our method on multiple benchmarks, including three image classification generic datasets (\ie, CIFAR 10/100~\cite{cifar} and ImageNet-100~\cite{deng09imagnet}), three fine-grained datasets from Semantic Shift Benchmark~\cite{vaze2022openset} (\ie, CUB~\cite{cub200}, Stanford Cars~\cite{StanfordCars} and FGVC-Aircraft~\cite{aircraft}), and three challenging datasets (\ie, Herbarium 19~\cite{herbarium}, ImageNet-R~\cite{inr} and ImageNet-1K~\cite{deng09imagnet}). 
We are the first to introduce ImageNet-R into the GCD task, which contains various renditions
of 200 ImageNet classes, thus challenging the GCD's assumption that the data comes from the same domain. The data splits are reported in \textit{Supp}.

\myPara{Evaluation and implementation details.} 
Following standard evaluation protocol in~\cite{vaze2022generalized,wen2023parametric}, we evaluate the performance with clustering accuracy (ACC). We use a CLIP~\cite{radford2021learning} pre-trained ViT-B/16~\cite{dosovitskiy2021an} as the image and text encoder. In the first stage, we train a fully connected layer. In the second stage, we remove the projector of the image encoder, resulting in 
features with a dimension of 768. Other details and the pseudo-code can be found in \textit{Supp}.

\subsection{Comparison with state of the arts}
In this section, we compare \OURS~with several
state-of-the-art methods.
GCD~\cite{vaze2022generalized} and SimGCD~\cite{wen2023parametric}  provide  paradigms for non-parametric and parametric   classification, thus we replace their backbone with CLIP  for a fair comparision,
denoted as GCD-CLIP  and SimGCD-CLIP.

\myPara{Evaluation on fine-grained and generic datasets.}
As shown in Tab.~\ref{subtab:ssb}, our method achieves consistently remarkable success on all three fine-grained datasets. Specifically, we surpass   SimGCD-CLIP by 5.3\%, 8.5\%, and 4.6\% on  `All' classes of CUB, Stanford Cars, and Aircraft, respectively. In fine-grained datasets, the visual conceptions of distinct classes exhibit high similarity, making it challenging for classification based solely on visual information. However, their text information can provide additional discriminative information. Consequently, our~\OURS~ significantly enhances classification accuracy through the reciprocal augmentation of text and visual information flows.
\begin{table}[t]

\begin{center}
\setlength{\tabcolsep}{2pt}
  \resizebox{\columnwidth}{!}{
\begin{tabular}{lccccccccc}
\toprule
&    \multicolumn{3}{c}{Herbarium 19} &
\multicolumn{3}{c}{ImageNet-1K} & \multicolumn{3}{c}{ImageNet-R} \\
\cmidrule(rl){2-4}\cmidrule(rl){5-7}\cmidrule(rl){8-10}
Method                                   & All  & Old  & New  & All  & Old  & New & All  & Old  & New\\
\midrule
$k$-means~\cite{macqueen1967some}  & 13.0 & 12.2 & 13.4 & - & - & - & - & - & -\\
RS+~\cite{Han2020automatically}          & 27.9 & {55.8} & 12.8 & - & - & -& - & - & - \\
UNO+~\cite{fini2021unified}               & 28.3 & {53.7} & 14.7 & - & - & - & - & - & -\\
ORCA~\cite{cao2022openworld}                     & 20.9 & 30.9 & 15.5 & - & - & -& - & - & - \\
$\mu$GCD~\cite{mugcd} & {45.8} & {61.9}  & 37.2& - & - & -& - & - & - \\
LegoGCD~\cite{legogcd} & 45.1& 57.4& {38.4} &62.4 &\textbf{79.5}& 53.8&- & - & -
\\
\midrule
GCD~\cite{vaze2022generalized}       & {35.4} & 51.0 & {27.0}& 52.5 & 72.5 & 42.2 & 32.5 & 58.0 & 18.9  \\
SimGCD~\cite{wen2023parametric}                      & {44.0} & {58.0} & {36.4}& {57.1} & {77.3} & {46.9} &29.5 &48.6 &19.4   \\ 

\midrule
GCD-CLIP       & {37.3} & 51.9 & {29.5}  & 55.0 &65.0 &50.0& 44.3 &  79.0& 25.8\\
SimGCD-CLIP                    & {48.9} & \textbf{64.7} & {40.3}& {61.0} & {73.1} & {54.9} & 54.9& 72.8&45.3 
\\
\midrule
\rowcolor{mygray}
\textbf{\OURS} (Ours)                    & \textbf{49.7} & {64.5} & \textbf{41.7}& \textbf{62.4} & {74.0} & \textbf{56.6} & \textbf{58.1} & \textbf{78.8} & \textbf{47.0}  \\
\bottomrule
\end{tabular}}
\vspace{-2mm}
\caption{Results (\%) on more challenging datasets.}\label{tab:herb19}
\end{center}

\vspace{-6mm}
\end{table}
In \cref{subtab:ssb}, we also present the performances
for three generic datasets.
Due to the low resolution of the CIFAR dataset and model biases (CLIP itself performs poorly on CIFAR100, with a zero-shot performance of 68.7), the results for novel classes are inferior compared to the DINO backbone. However, despite the inherent limitations in the discriminative capability of CLIP itself,
our method still achieves an improvement of 0.4\% on `Old' classes of CIFAR10 and 2.2\% on `New' classes of CIFAR100, compared to SimGCD-CLIP.
For ImageNet-100, SimGCD-CLIP has achieved an exceptionally saturated result of  90.8\% on `All' classes, further advancements pose considerable challenges. 
However, leveraging the additional modality information, \OURS~elevates the performance ceiling to an impressive 91.7\%.

\myPara{Evaluation on more challenging datasets.} As shown in \cref{tab:herb19}, \OURS~outperforms all other methods for both `All' and `New' classes on Herb19 and ImageNet-1K datasets. In particular, our method achieves a notable improvement of 1.4\% and 1.7\% on `New' classes of Herb19 and ImageNet-1K, respectively.
Furthermore, the suboptimal performance of GCD and SimGCD with the DINO backbone on the ImageNet-R dataset highlights the difficulty of DINO in discovering new categories with multiple domains.
Despite multiple domains for images of the same category, their textual information remains consistent. 
Our method effectively integrates text information, resulting in a substantial improvement of 3.2\% and 6.0\% over the sota for `All' classes and `Old' classes, respectively.
It is worth noting that, owing to the text consistency within the same category,
our text branch 
achieves remarkable 62.6\%   and 63.5\% accuracy for
`All' classes of Imagenet-R and ImageNet-1K, respectively.

\begin{table}[t]

\begin{center}
\setlength{\tabcolsep}{2.5pt}
  \resizebox{\columnwidth}{!}{
  \begin{tabular}{lccccccccc} \toprule
    \multirow{2}{*}{} & \multirow{2}{*}{TES} & \multirow{2}{*}{Dual-branch} & \multirow{2}{*}{CICO} & \multicolumn{3}{c}{Stanford Cars} & \multicolumn{3}{c}{CIFAR100} \\ 
    \cmidrule(rl){5-7} \cmidrule(rl){8-10}
    & & & & \multicolumn{1}{c}{All}  & \multicolumn{1}{c}{Old}  & New & \multicolumn{1}{c}{All} & \multicolumn{1}{c}{Old}  & New \\ \midrule
    \multicolumn{1}{c}{(1)} & \xmark & \xmark & \xmark & \multicolumn{1}{c}{70.0} & \multicolumn{1}{c}{83.4} & 63.5 & \multicolumn{1}{c}{81.1} & \multicolumn{1}{c}{85.0} & 73.3 \\ 
    \multicolumn{1}{c}{(2)} & \cmark & \cmark & \xmark & \multicolumn{1}{c}{76.2} & \multicolumn{1}{c}{85.3} & 71.7 & \multicolumn{1}{c}{81.0} & \multicolumn{1}{c}{85.3} & 72.3 \\ 
    \multicolumn{1}{c}{(3)} & \cmark & \cmark & \cmark & \multicolumn{1}{c}{\textbf{78.5}} & \multicolumn{1}{c}{\textbf{86.8}} & \textbf{74.5} & \multicolumn{1}{c}{\textbf{82.1}} & \multicolumn{1}{c}{\textbf{85.5}} & \textbf{75.5} \\
    \bottomrule
\end{tabular}}
\vspace{-2mm}
\caption{Ablation study of different components.}
\label{tab:diff_com}
\end{center}
\vspace{-9mm}
\end{table}

\subsection{Ablation study and analysis}
\label{ssec:ablation}

\myPara{Effectiveness of different components.}
To evaluate the
effectiveness of different components, we conduct an ablation study on SCars and CIFAR100 in Tab.~\ref{tab:diff_com}. Comparing (2) with (1), leverage the text features generated by TES, resulting in a 6.2\% improvement on SCars' `All' classes and a 0.3\% improvement on CIFAR100's `Old' classes. Furthermore, comparing (3) with (1), CICO enables the two branches to exchange information and mutually benefit from each other, resulting in remarkable improvements of 11\% on SCars' `New' and 2.2\% on CIFAR100's `New'.

\myPara{Comparison with different fusion methods.}
In Tab.~\ref{tab:diff_fu}, we compare our dual-branch strategy with other modality fusion methods, including concatenation and mean. 
Although they may show improvements due to the  multi-modal information, we demonstrate that  joint learning of the two branches is more effective as it encourages  the model to have complementary and aligned discriminative capabilities for visual and text features of the same
class, leading to more discriminative multi-modal prototypes.
\begin{table}[t]
\begin{center}
\setlength{\tabcolsep}{2pt}
  \resizebox{\columnwidth}{!}{

\begin{tabular}{lccccccccc}
\toprule
\multirow{2}{*}{}        & \multirow{2}{*}{Dual-branch} & \multirow{2}{*}{Concat} & \multirow{2}{*}{Mean} & \multicolumn{3}{c}{Stanford Cars}                                                          & \multicolumn{3}{c}{CIFAR100}                                                       \\ 
\cmidrule(rl){5-7}
\cmidrule(rl){8-10}
                                           &                              &                                   &                                     & \multicolumn{1}{c}{All}  & \multicolumn{1}{c}{Old}  & New                      & \multicolumn{1}{c}{All}  & \multicolumn{1}{c}{Old}  & New                            \\ 
\midrule

\multicolumn{1}{c}{(1)}     & \xmark                          & \cmark                                & \xmark                                  & \multicolumn{1}{c}{68.9} & \multicolumn{1}{c}{79.1} & 64.0                           & \multicolumn{1}{c}{79.9} & \multicolumn{1}{c}{85.5} & 68.7                          \\ 

\multicolumn{1}{c}{(2)}       & \xmark                           & \xmark                               & \cmark                                  & \multicolumn{1}{c}{72.0} & \multicolumn{1}{c}{85.0} & 65.6                           & \multicolumn{1}{c}{81.1} & \multicolumn{1}{c}{84.3} & 74.8                           \\


\multicolumn{1}{c}{(3)}     & \cmark                          & \xmark                               & \xmark                                  & \multicolumn{1}{c}{\textbf{78.5}} & \multicolumn{1}{c}{\textbf{86.8}} & \textbf{74.5} & \multicolumn{1}{c}{\textbf{82.1}} & \multicolumn{1}{c}{\textbf{85.5}} & \textbf{75.5}                          \\
 
\bottomrule
\end{tabular}}
\end{center}
\vspace{-5mm}
\caption{Comparison with different fusion methods.}
\label{tab:diff_fu}
\vspace{-3mm}
\end{table}

\begin{table}[t]
\begin{center}
\setlength{\tabcolsep}{2pt}
\resizebox{\columnwidth}{!}{
\begin{tabular}{llcccc}
\toprule
\multicolumn{2}{c}{Method}  &  Total Params & All & Old & New \\ 
\midrule
 \textit{Baseline}
 &SimGCD-CLIP & 92.2M & 71.7 & 76.5 & 69.4\\
\midrule
 {\multirow{2}{*}{{\textit{Text-Retrieval }}}}
 &WordNet & 155.8M & 69.8	& 77.1	& 66.2\\
 &CC3M & 155.8M & 72.3 &	\textbf{79.1}	 & 68.9 \\
 \midrule
 {\multirow{2}{*}{{\textit{VQA}}}} & BLIP (ViT-L) &625.8M&  67.1 & 	74.3  &	63.5 \\
 & BLIP-2 (opt2.7b)  & 3.9B&71.3&	73.5	&70.2 \\
 \midrule
  {\multirow{2}{*}{{\textit{Captioning}}}} & BLIP (ViT-L) & 625.8M& 40.5 & 	54.6  &	33.4 \\
 & BLIP-2 (opt2.7b)& 3.9B  &42.6&	56.1	&35.8 \\
 \midrule
   {\multirow{3}{*}{{\textit{Feature-Generation }}}} & {TES (Ours)} & 165.1M& \textbf{77.0} 	&78.1	 &\textbf{76.4} \\
 & {TES {w/o}} $\mathcal{L}_{align}$ & 165.1M& 74.7  & 	76.9	& 73.5 \\
 & {TES {w/o}} $\mathcal{L}_{distill}$ &  165.1M& 75.3 	& 77.5	&  74.2\\
\bottomrule
\end{tabular}}
\end{center}
\vspace{-5mm}
\caption{Experiments on different pseudo text embeddings.}
\label{tab:diff_tes}
\vspace{-6mm}
\end{table}

\myPara{Effectiveness of TES.}
To demonstrate the superiority of TES, we conduct experiments that replaced the embeddings generated by TES with embeddings obtained through \textit{Text-Retrieval}, \textit{VQA}, and \textit{Captioning} on the CUB dataset. 
For \textit{Text-Retrieval}, we retrieve the most similar text for each image from two corpora (WordNet~\cite{wordnet} and CC3M~\cite{sharma2018conceptual}) based on the cosine similarity between the image and text embeddings. 
We use BLIP~\cite{li2022blip} and BLIP-2~\cite{li2023blip2} to perform the \textit{Captioning} and use the question ``What's the name of the bird in the image?'' to perform \textit{VQA}.
As in \cref{tab:diff_tes}, due to the high visual similarity in fine-grained images, category names retrieved or generated through VQA are often imprecise, making them less effective.
Meanwhile, captioning methods tend to describe object poses and scenes rather than class-specific information, leading to varied captions for samples of the same class, which significantly harms category discovery.
Our method achieves the best performance with moderate parameters.
We provide additional experiments about TES in \textit{Supp}, including its architectural design, feature distribution, and flexibility.

\myPara{Results using different prompts.}
In our method, we use a simple prompt: ``a photo of a \{CLS\}."~\cref{tab:prompt} presents experimental results exploring the use of alternative prompts: (1) ``a photo of a \{CLS\},"  (2) ``a photo of a \{CLS\}, which is a type of bird/car," (3) descriptions of \{CLS\} generated by LLM (GPT4o-mini), and (4) average the textual features of (1) to (3).  Our prompt is simple yet effective, while more finely designed prompts can further improve performance.

\begin{table}[t]

\centering
\setlength{\tabcolsep}{9pt}
\resizebox{\columnwidth}{!}{
\begin{tabular}{ccccccc}
\toprule

\multirow{2}{*}{Prompts} & \multicolumn{3}{c}{CUB} & \multicolumn{3}{c}{Stanford Cars} \\ 
\cmidrule(rl){2-4}
\cmidrule(rl){5-7}
                                & \multicolumn{1}{c}{All}  & \multicolumn{1}{c}{Old}  & New   & \multicolumn{1}{c}{All}  & \multicolumn{1}{c}{Old}  & New \\

\midrule
(1) & 77.0&78.1&76.4& 78.5&86.8&\textbf{74.5}\\
(2) & 76.3&78.2&75.4& 78.5 &88.2 &73.8 \\
(3) & 76.8&\textbf{78.7}&75.8&78.6 &\textbf{90.4} & 72.9 \\
(4) & \textbf{78.3}&77.6&\textbf{78.7}& \textbf{79.1} & 88.8 & 74.3\\
\bottomrule
\end{tabular}}
%
 \vspace{-2mm}

\caption{Results using different prompts.}
\vspace{-3mm}
\label{tab:prompt}

\end{table}

\begin{table}[t]

\centering
\setlength{\tabcolsep}{6pt}
\resizebox{\columnwidth}{!}{
\begin{tabular}{lcccccc}
\toprule

\multirow{2}{*}{Methods} & \multicolumn{3}{c}{NEV} & \multicolumn{3}{c}{TV-100} \\ 
\cmidrule(rl){2-4}
\cmidrule(rl){5-7}
                                & \multicolumn{1}{c}{All}  & \multicolumn{1}{c}{Old}  & New   & \multicolumn{1}{c}{All}  & \multicolumn{1}{c}{Old}  & New \\

\midrule
CLIP(zero-shot) & 10.7 &-&- & 1.93&-&-\\
SimGCD & 54.7 & 88.0  & 38.0 & 35.2 &50.3& 29.2\\
SimGCD-CLIP        & 79.1 & \textbf{96.7}  & 70.3 & 55.7 &75.8 &47.8 \\
\textbf{\OURS} (Ours)      & \textbf{85.3 }  &  96.0   &  \textbf{80.0 }&  \textbf{57.1 }&\textbf{77.3} &\textbf{49.2}\\
\bottomrule
\end{tabular}}
%
 \vspace{-2mm}

\caption{The results on the NEV and TV-100 datasets.}
\vspace{-6mm}
\label{tab:g_v}

\end{table}

\myPara{Discussion about using CLIP in GCD.}
A key purpose of GCD is to discover novel classes, which highly rely on the initial representation discrimination provided by the backbone model.
Due to the strong generalization ability of CLIP, it can encode more discriminative features, making it a natural idea to introduce CLIP into the GCD task.
One concern is that CLIP may have seen unknown classes or class names in GCD. 
Therefore, we discuss the significance of using CLIP in GCD from three perspectives.
\textbf{Methodological Significance:} Even CLIP is pre-trained on a vast dataset with potential overlap, its knowledge is implicit and unstructured. Effectively using it for GCD demands novel methodologies, particularly in exploring how to use the text encoder for unlabelled data. The improvements of our method over SimGCD-CLIP validate this significance. \textbf{Forward-looking Significance:}
To evaluate the ability of category discovery methods in scenarios unseen by CLIP, we constructed a small fine-grained dataset of new energy vehicles (NEV) introduced in 2023, where CLIP lacks prior knowledge.
As shown in \cref{tab:g_v}, experiments on NEV and TV-100~\cite{zhou2024tv}(a TV series dataset that the pre-trained CLIP model has not been exposed to) demonstrate that even for categories not seen by CLIP, leveraging the text modality remains crucial for effective category discovery. This serves as a forward-looking exploration of how CLIP’s generalization can address future GCD tasks involving truly novel data.
\textbf{Practical Implications:} Exploring CLIP's potential in realistic scenarios is meaningful, thus, we present experiments on medical and ultra-fine-grained datasets in \textit{SSSupp}. Our work lays the foundation for leveraging CLIP to address challenging GCD applications.

\subsection{Qualitative results}

\begin{figure}[ht]
\vspace{-3mm}
    \centering
\includegraphics[width=0.95\columnwidth]{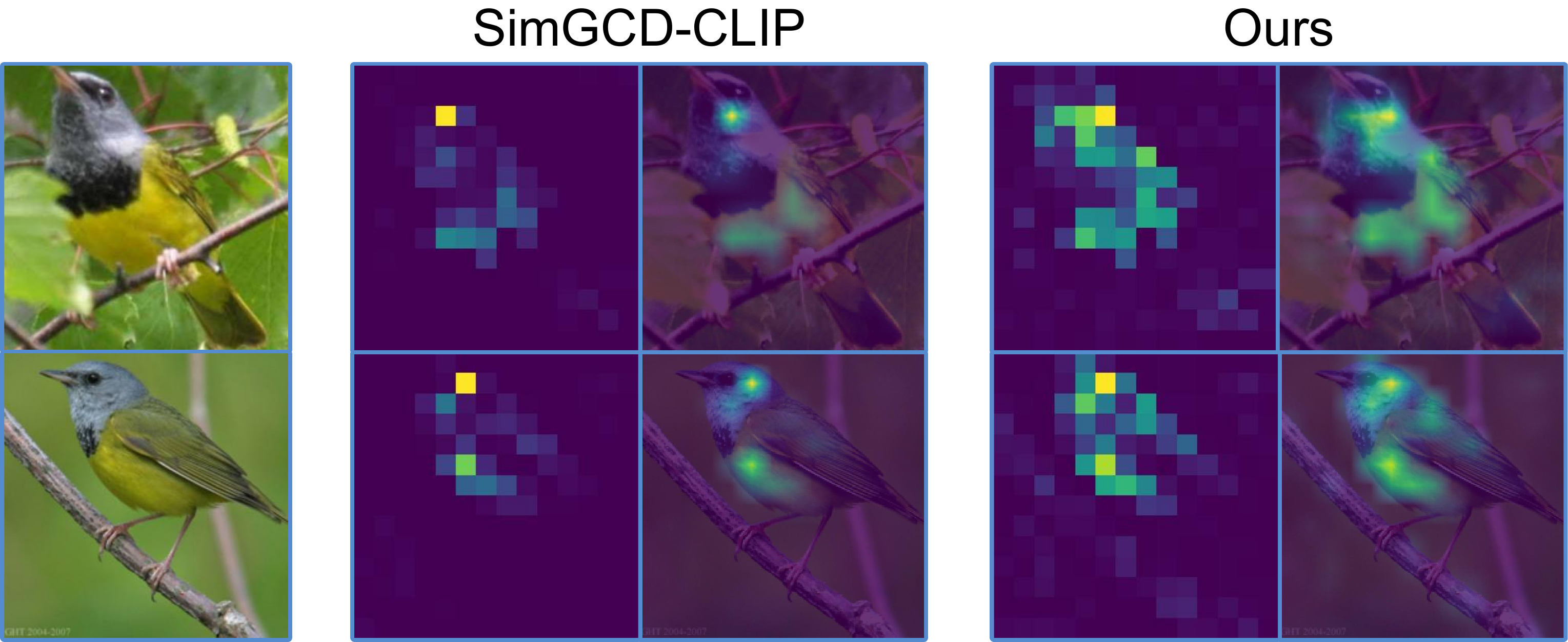}
    \vspace{-3mm}
    \caption{Attention map visualization of class tokens.
    }
    \label{fig:cub_ours_attnmp}
    \vspace{-4mm}
\end{figure}

\myPara{Attention map visualization.} As in~\cref{fig:cub_ours_attnmp}, compared to SimGCD-CLIP, our method additionally focuses on the feather texture of birds, which is crucial for distinguishing visually similar bird species. With the assistance of text information, the attention maps of our visual branch become more refined, focusing on more discriminative regions.

\begin{figure}[ht]
    \centering
\vspace{-6mm}
    \includegraphics[width=\columnwidth]{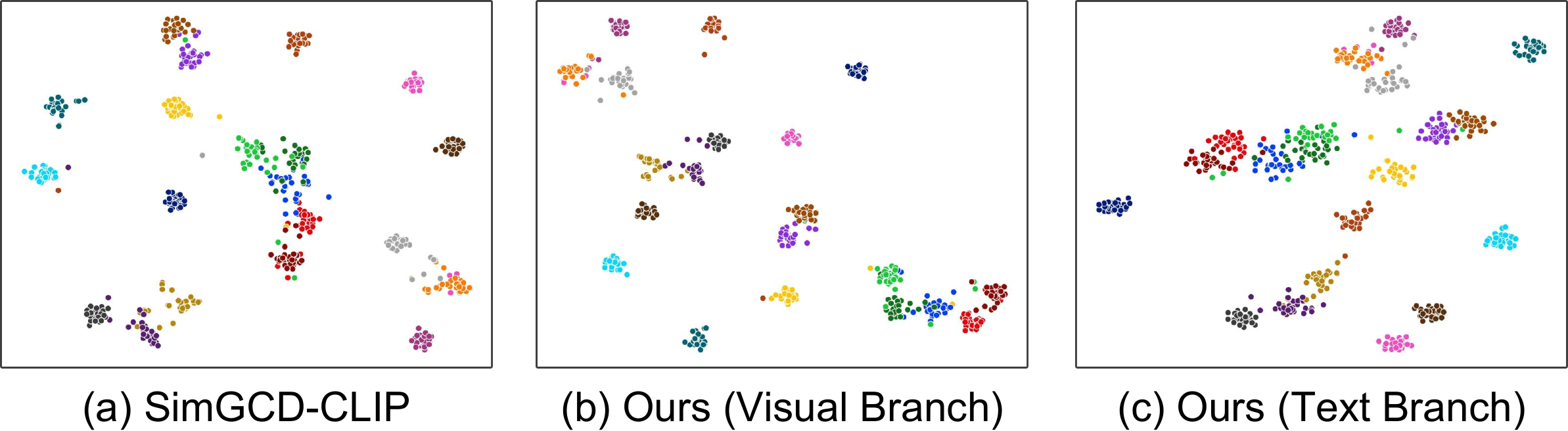}
\vspace{-6mm}
    \caption{The t-SNE visualizations.
    }
    \label{fig:cub_tsne}
    \vspace{-6mm}
\end{figure}

\myPara{The t-SNE visualization.}
\cref{fig:cub_tsne} shows the t-SNE visualization of visual and text features on the randomly sampled 20 classes of the CUB dataset. Both the visual and text features of our method exhibit clearer and compacter clusters. We provide more visualizations and cluster results in \textit{Supp}.

%% file: sec/6_conclusion.tex
\section{Conclusions}
In this work,
we propose to leverage multi-modal information
to solve the GCD task. In particular, we introduce a text embedding synthesizer to generate
pseudo text embeddings for unlabelled data. 
Our text embedding synthesizer module makes it possible to use CLIP's text encoder, thus unlocking the multi-modal potential for the GCD task.
Meanwhile, we use a dual-branch training strategy with a cross-modal instance consistency objective, which facilitates collaborative action and mutual learning between the different modalities.
Our research extending the GCD to a multi-modal paradigm and the superior performance on multiple benchmarks demonstrates the effectiveness of our method.

\myPara{Limitations and future work.}
A limitation of our approach is
that we treat visual and text information as equally important. In fact, some
samples may have richer and more discriminative visual information than textual
information, and vice versa. A more appropriate approach might involve enabling
the model to adaptively leverage multimodal information, autonomously assessing which modality’s information is more crucial. We will delve deeper into this
aspect in our future work.

%% file: sec/7_ack.tex
\section*{Acknowledgement}
This work is funded by  
Shenzhen Science and Technology Program (JCYJ20240813114237048),
NSFC (NO. 62206135, 62225604),
Young Elite Scientists Sponsorship Program by CAST (2023QNRC001),
the Fundamental Research Funds for the Central Universities 
(Nankai Universitiy, 070-63233085),
and ``Science and Technology Yongjiang 2035" key technology breakthrough plan project (2024Z120).
Computation is supported by the Supercomputing Center of Nankai University.

%% file: sec/X_suppl.tex
\maketitlesupplementary

\setcounter{section}{7}
\setcounter{figure}{4}
\setcounter{table}{7}
\setcounter{equation}{12}

This Supplementary Material provides the following  sections: 
\begin{itemize}
\item  Experimental setup (\cref{supp1})
\item  Discussion about Using CLIP in GCD (\cref{disscusion})
\item  Additional experiments and analysis (\cref{sA})
\item  Pseudo-code (\cref{ps})
\item Different pre-trained models (\cref{ssec:pm})
\item Cluster results of GET (\cref{sB})
\item Limitations and broader impact (\cref{lm})
\end{itemize}

\section{Experimental setup}
\label{supp1}
\myPara{Datasets.}We evaluate our method on multiple benchmarks, including three image classification generic datasets (\ie, CIFAR 10/100~\cite{cifar} and ImageNet-100~\cite{deng09imagnet}), three fine-grained datasets from Semantic Shift Benchmark~\cite{vaze2022openset} (\ie, CUB~\cite{cub200}, Stanford Cars~\cite{StanfordCars} and FGVC-Aircraft~\cite{aircraft}), and three challenging datasets (\ie, Herbarium 19~\cite{herbarium}, ImageNet-R~\cite{inr} and ImageNet-1K~\cite{deng09imagnet}). 
Notice that we are the first to introduce ImageNet-R into the GCD task, which contains various renditions
of 200 ImageNet classes, thus challenging the GCD's assumption that the data comes from the same domain.
For ImageNet-R, we subsample the first 100 classes as old classes, leaving the rest as new classes; the labeled dataset $\mathcal{D}_l$ consists of half of the old class samples, while the other half and all the new class samples are used to construct unlabelled dataset $\mathcal{D}_u$. 
Furthermore, we conduct experiments on the TV100 dataset~\cite{zhou2024tv}, {a TV series dataset that the pre-trained CLIP model has not been exposed to}. We use the first 50 classes as old categories and the remaining 50 classes as new categories.
As for other benchmarks,
we follow the previous~\cite{vaze2022generalized, wen2023parametric} to sample $\mathcal{D}_l$ and $\mathcal{D}_u$.
The details of the standard datasets we evaluate on are
shown in \cref{tab:dataset}.
\begin{table}[h]   
  \small
  \centering
  \resizebox{.98\columnwidth}{!}{
  \begin{tabular}{lrcrc} \toprule
    & \multicolumn{2}{c}{Labelled}  & \multicolumn{2}{c}{Unlabelled}\\
    \cmidrule(rl){2-3}\cmidrule(rl){4-5}
    Dataset          & Images   & Classes   & Images   & Classes \\ \midrule
    CIFAR10~\cite{cifar} & 12.5K     & 5         & 37.5K     & 10 \\
    CIFAR100~\cite{cifar} & 20.0K     & 80        & 30.0K     & 100 \\
    ImageNet-100~\cite{deng09imagnet}       & 31.9K     & 50        & 95.3K     & 100 \\
    CUB~\cite{cub200} & 1.5K      & 100       & 4.5K      & 200 \\
    Stanford Cars~\cite{StanfordCars}      & 2.0K      & 98        & 6.1K      & 196 \\    
    FGVC-Aircraft~\cite{aircraft}      & 1.7K      & 50        & 5.0K      & 100 \\
    Herbarium 19~\cite{herbarium}       & 8.9K      & 341       & 25.4K     & 683 \\
    ImageNet-R~\cite{inr}       & 7.7K     & 100        & 22.3K     & 200 \\
    ImageNet-1K~\cite{deng09imagnet}       & 321K     & 500        & 960K     & 1000 \\
    \bottomrule
  \end{tabular}}
 \caption{The details of the standard datasets we evaluate on.}
\label{tab:dataset}
\end{table}

\myPara{The NEV dataset}
As mentioned in the main paper, we conduct a toy experiment to prove that our TES can deal with a scenario where CLIP lacks information on a specific category class. Since CLIP saw most of the visual concepts and corresponding texts before 2022,  we constructed a small dataset of new energy vehicles (NEV) that appeared in 2023. As  in \cref{tab:gv_name}, the NEV dataset contains 12 categories, each with 50 images from the Internet, and the classnames of the dataset consist of the brand and model of the car. We split them in the same way as standard benchmarks. 
\begin{table}[h]

\vspace{-1mm}
\centering
\resizebox{\columnwidth}{!}{
\begin{tabular}{ll}
\toprule
Old classes & New classes\\ \midrule
    BMW\_xDrive\_M60 &Geely\_Jiyue\_01\\  BYD\_Seagull & Geely\_Zeeker\_X\\  BYD\_Song\_L &Mercedes-Benz\_EQE\_SUV\\ BYD\_Yangwang\_U8 & SAIC-Motor\_MG\_Cyberster \\  GAC-Motor\_Trumpchi\_ES9 &SAIC-Motor\_Rising\_F7\\  Geely\_Galaxy\_E8&XPeng\_X9 \\  
\bottomrule
\end{tabular}}
%
\vspace{-3mm}
\caption{The class names for the NEV dataset.}
\vspace{-4mm}
\label{tab:gv_name}
\end{table}

\myPara{Implementation details.}
We use a CLIP~\cite{radford2021learning} pre-trained ViT-B/16~\cite{dosovitskiy2021an} as the image and text encoder. 
In the first stage, we train a fully connected layer to transfer image embeddings into pseudo-tokens. 
In the second stage, the projector of the image encoder is removed, resulting in features with a dimension of 768.
The exception is ImageNet-1K, we remain and fine-tune the last projection layer, which avoids gradient explosion and improves results with lower computational cost,
resulting in features with a dimension of 512. 
We use a single linear layer to turn pseudo text embeddings generated by TES into learnable embeddings while changing their dimensions (512 to 768) to match those of the visual features.
The batch size is fixed to 128 for training and 256 for testing. 
Training is done with an SGD optimizer and an initial learning rate of 0.1 decayed by a cosine annealing rule.
We train for 200 epochs on each dataset in both two stages. 
In the first stage, we set the number of pseudo text tokens to 7.
The balance coefficient $\lambda$  is set to 0.35 as~\cite{vaze2022generalized}, and $\lambda_c$ is set to 1. 
The temperature value $\tau_a$ is set to 0.01 while other temperature values 
$\tau_{sc}$, $\tau_c$,  $\tau_s$ , $\tau_t$ and the balanced value $\epsilon$ are as same as \cite{wen2023parametric}.
The augmentation exactly follows the previous, in which RandomCrop creates two views.
All experiments are conducted with 4 NVIDIA GeForce RTX 3090 GPUs.

\begin{table}[t]
\small
    \centering
\begin{tabular}{lcccc}
\toprule
\multirow{2}[3]{*}{Method}   &\multirow{2}[3]{*}{Backbone}                               &\multicolumn{3}{c}{ NCT-CRC-HE }\\
 \cmidrule(lr){3-5} 
& & All  & Old  & New   \\
\midrule
SimGCD &DINO&77.1&79.9&75.1\\
SimGCD &CLIP&79.1&93.2&69.2\\
\rowcolor{mygray}
\textbf{GET (ours)} & CLIP & \textbf{83.8} & \textbf{94.5} & \textbf{76.3}

\\
\bottomrule
\end{tabular}
\vspace{-2mm}
\caption{Results on  the medical dataset.}
\label{tab:med}
\vspace{-7mm}
\end{table}

\section{Discussion about Using CLIP in GCD}
\label{disscusion}
An evident fact is that using a more powerful backbone facilitates the transfer of knowledge learned from labeled data to unlabeled data~\cite{vaze2022generalized,wen2023parametric,mugcd,sptnet}. Due to the strong generalization ability of CLIP, it can encode more discriminative features, and its multi-modal information aids in discovering new categories, making it a natural choice for introducing CLIP. 
As discussed in the main text, using CLIP in GCD has three significances: methodological significance, forward-looking significance, and practical implications. 

We argue that \textbf{the key to leveraging CLIP for GCD lies in how to use its text encoder}, given the presence of unlabeled data in GCD tasks. 
In this section, we provide supplementary analyses to complement the discussions in the main text.
To be specific, we validate the effectiveness of our method and substantiate the incorporation of CLIP into GCD by addressing the following questions:

\myPara{1. Does the performance gain originate from the CLIP (text encoder) being pre-exposed to the new categories?}

In other words, we need to verify the effectiveness of our method on categories that are unseen by CLIP.  
The experimental results on the NEV and TV-100 datasets (Tab. \textcolor{red}{7} in the main paper) demonstrate the effectiveness of our method in scenarios where CLIP lacks prior information.

The intuition behind our TES can be explained from two perspectives. First, our trained TES can be considered as a special fine-tuned text encoder. This text encoder takes visual images as input and produces corresponding textual features as output. Our align loss 
 ensures modal alignment, while the distill loss facilitates the model's adaptation to the dataset's distribution. Second,  TES can be regarded as a caption model~\cite{merullo2022linearly}. For each input image, TES assigns a corresponding caption, expressing each caption in the form of modal-aligned text features.
The text embeddings or captions corresponding to images can serve as valuable supplementary information, assisting the GCD task in a multimodal manner.

\myPara{2. In more realistic scenarios where class names (either old or new) cannot be generated or retrieved, does our method remain effective?}

To address this concern, we first conducted GCD experiments on a medical dataset: the NCT-CRC-HE dataset. 
The NCT-CRC-HE~\cite{kather_jakob_nikolas_2018_1214456} dataset comprises histological images of human colorectal cancer, containing nine categories. We selected the first five categories as the old classes.  
For the medical dataset, generating or retrieving new class names is challenging, its class names need expert knowledge. Our method directly generates text features aligned with visual features, and the experimental results in~\cref{tab:med} demonstrate its effectiveness.

Moreover, in certain scenarios, it is also difficult to obtain the class names of base categories. For example, in ultra-fine-grained datasets~\cite{Liu_2024_CVPR,yu2021benchmark}, different categories represent different types of soybean leaves. In such cases, category discovery using CLIP becomes significantly more challenging. To address this, we remove the distillation loss in TES, allowing the use of CLIP's text encoder even when base class names are unavailable. \cref{tab:gcd_uf} presents the experimental results on three ultra-fine-grained datasets, demonstrating the effectiveness of our approach. 
Meanwhile, the results in Tab. \textcolor{red}{5} of the main paper further demonstrate the effectiveness of our TES in scenarios where the class names of base categories are unavailable, by removing the distillation loss.

\begin{table}[t]
\small
\centering
\setlength{\tabcolsep}{2pt}
\resizebox{\columnwidth}{!}{%
    \begin{tabular}{lccc|ccc|ccc}
      \toprule
      \multirow{2}{*}{}  & \multicolumn{3}{c}{SoyAgeing-R1} & \multicolumn{3}{c}{SoyAgeing-R3} & \multicolumn{3}{c}{SoyAgeing-R4}  \\
\cmidrule(rl){2-4}\cmidrule(rl){5-7}\cmidrule(rl){8-10} 
  Method    & All  & Old  & New  & All  & Old  & New  &  All & Old  & New    \\
      \midrule
      SimGCD& 37.3 & 48.4&31.7 &32.7&47.2&25.5&35.4&46.4&29.9 \\
      \rowcolor{mygray}
      \textbf{GET (ours)} & \textbf{47.9} &\textbf{56.5}&\textbf{43.6}&\textbf{46.0}&\textbf{55.2}&\textbf{41.4}&\textbf{46.6}&\textbf{52.8}&\textbf{43.4} \\
        
      \bottomrule
    \end{tabular}
}
\vspace{-2mm}
\caption{Results on ultra-fine-grained datasets using CLIP backbone.}
\label{tab:gcd_uf}
\vspace{-7mm}
\end{table}

\section{Additional Experiments and Analysis}
\label{sA}

\begin{figure}[ht]
	\centering
	\includegraphics[width=.8\columnwidth]{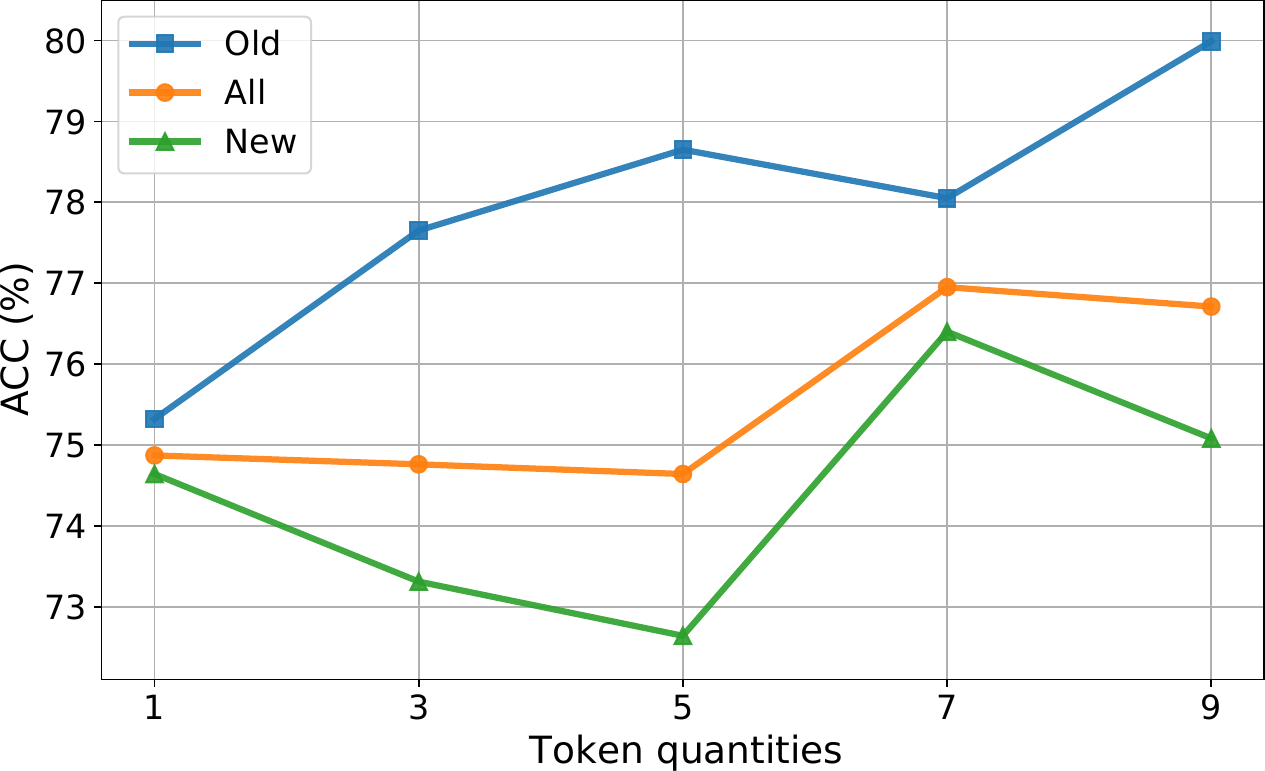}
 \\ \vspace{0.8em}
 \includegraphics[width=.8\columnwidth]{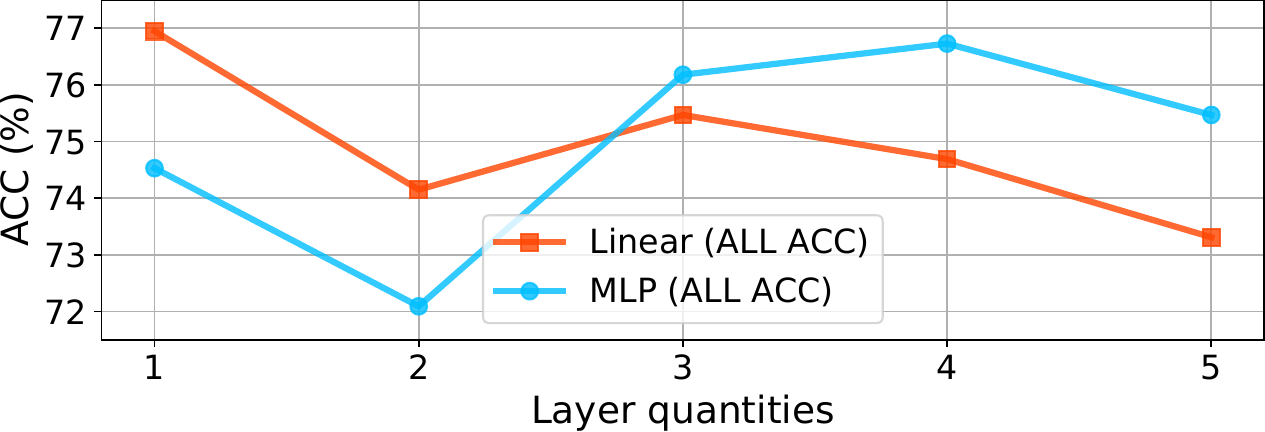}
 \vspace{-2mm}
        \caption{Experiments on the pseudo-tokens 
         and layers in TES.}
        \label{fig:tokens}
    \vspace{-4mm}
\end{figure}

\myPara{The architecture of TES.}
In TES, we use a single linear layer to  transform
the visual embedding to pseudo tokens and set the number of pseudo text tokens to 7 across all datasets. Experiment results on the architecture of TES for the CUB dataset are presented in~\cref{fig:tokens}, proving that a single linear layer can effectively transfer visual features into text tokens while reducing the computational cost.

\begin{figure}[h]
\vspace{-3mm}
    \centering
    \includegraphics[width=\columnwidth]{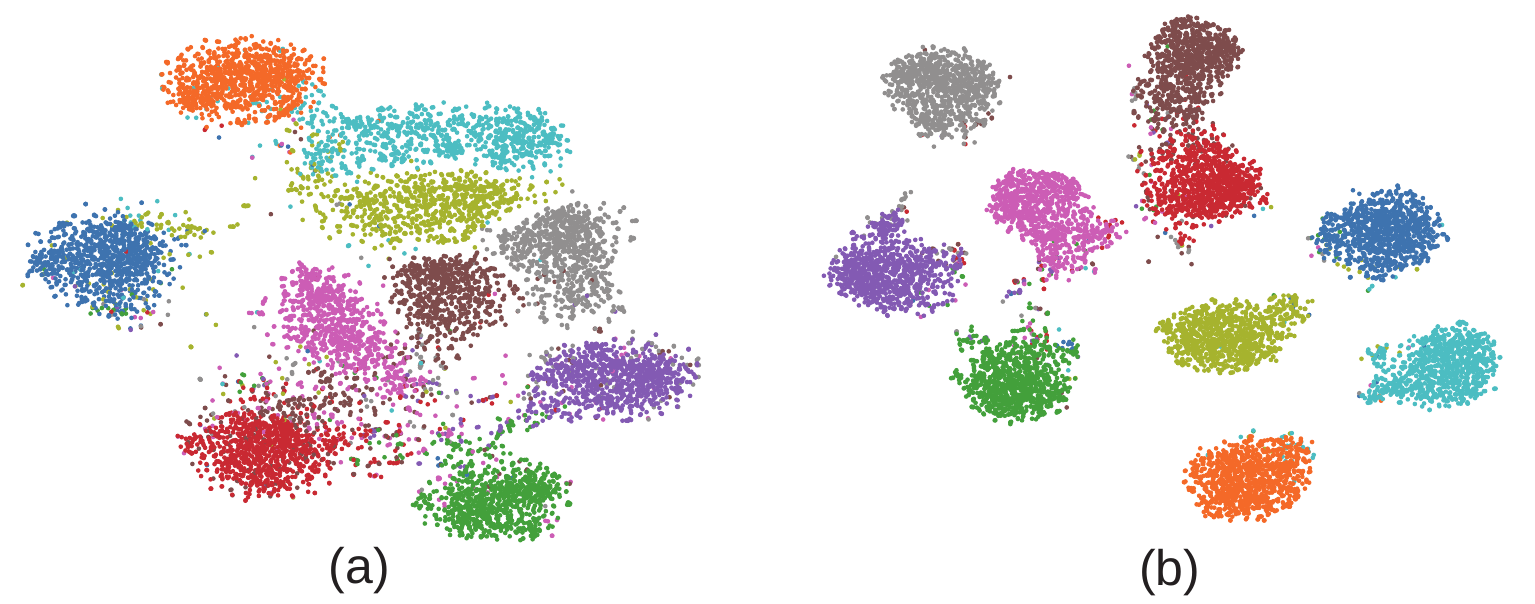}
    \caption{t-SNE visualization of text features for all classes on CIFAR10 test set. (a) shows the distribution of text features generated by TES, while (b) shows the learnable text features.
    }
    \label{fig:c10_tsne}
    \vspace{-2mm}
\end{figure}

\myPara{Effectiveness of text embedding synthesizer.}
In order to prove that our text embedding synthesizer can generate reliable and discriminative representations, we visualize the text embeddings of CIFAR10 with t-SNE. As shown in Fig.~\ref{fig:c10_tsne}, the initial text embeddings within the same class exhibit clear clustering, and the learnable embeddings further produce compacter clusters. Moreover, we introduce TES into the non-parametric GCD by straightforwardly concatenating text and image features before semi-supervised k-means classification. As in Tab.~\ref{subtab:gcd+tes}, with the help of text information, GCD gains about 5\% average improvement on `All' classes over 3 datasets, demonstrating the importance of multi-modal information in GCD task and our TES can be widely used in multiple GCD methods. 

\begin{table*}[!t]
\centering

  \setlength{\tabcolsep}{3pt} 
\resizebox{\textwidth}{!}{%
\begin{tabular}{lcccccccccccccccccc}
\toprule
&   \multicolumn{3}{c}{CUB} & \multicolumn{3}{c}{Stanford Cars} & \multicolumn{3}{c}{FGVC-Aircraft} &    \multicolumn{3}{c}{CIFAR10} & \multicolumn{3}{c}{CIFAR100} & \multicolumn{3}{c}{ImageNet-100} \\
\cmidrule(rl){2-4}\cmidrule(rl){5-7}\cmidrule(rl){8-10}\cmidrule(rl){11-13}\cmidrule(rl){14-16}\cmidrule(rl){17-19}
Methods        & All  & Old  & New  & All  & Old  & New  & All  & Old  & New  & All  & Old  & New  & All  & Old  & New  & All  & Old  & New\\
\midrule
PromptCAL~\cite{zhang2023promptcal} & {62.9} & {64.4} & {62.1} & {50.2} & 70.1 & {40.6} & {52.2} &52.2 & {52.3} & \textbf{97.9} & {96.6} & {98.5} & {81.2} & 84.2 & {75.3} & {83.1} &92.7 & {78.3}\\
PromptCAL-CLIP & {65.5} & {68.7} & {63.9} & {74.0} & 80.8 & {70.8} & {54.5} &61.8 & {51.0}& {88.7} & {96.5} & {84.8} & {80.5} & 82.4 & \textbf{76.8} & {87.4} &93.6 & {84.3}  \\
\midrule
\rowcolor{mygray}
\textbf{\OURS} (Ours)                    & \textbf{77.0} & \textbf{78.1} & \textbf{76.4} & \textbf{78.5} & \textbf{86.8} & \textbf{74.5} & \textbf{58.9} & \textbf{59.6} & \textbf{58.5} & {97.2} & {94.6} & \textbf{98.5} & \textbf{82.1} & \textbf{85.5} & {75.5} & \textbf{91.7} & \textbf{95.7} & \textbf{89.7}  \\

\bottomrule
\end{tabular}}
\caption{Results of PromptCAL-CLIP.} 
\label{subtab:p_ssb}
\vspace{-5mm}
\end{table*}

\begin{table}[t]
\begin{center}
\setlength{\tabcolsep}{2.5pt}

  \resizebox{\columnwidth}{!}{
  \begin{tabular}{lccccccccc} \toprule
    & \multicolumn{3}{c}{FGVC-Aircraft} & \multicolumn{3}{c}{ImageNet-100} & \multicolumn{3}{c}{ImageNet-R} \\
    \cmidrule(rl){2-4}\cmidrule(rl){5-7}\cmidrule(rl){8-10}
    Method  & All & Old & New & All & Old & New & All & Old & New \\ \midrule
    GCD-CLIP & {45.3 } & {44.4 } & {45.8} & {75.8 } & 87.3  & {70.0} & {44.3 } & 79.0  & 25.8 \\
    +TES & \textbf{49.6} & \textbf{49.3} & \textbf{49.8} & \textbf{80.0} & \textbf{95.1} & \textbf{72.4} & \textbf{49.4} & \textbf{79.4} & \textbf{33.5} \\ \bottomrule
  \end{tabular}}
  \end{center}
  
  \vspace{-4mm}
  \caption{Effectiveness of TES  in non-parametric GCD.}
  \label{subtab:gcd+tes} 
  \vspace{-5mm}
\end{table}

\begin{figure}[t]
    \centering
\setlength{\tabcolsep}{2pt}
    \includegraphics[width=\columnwidth]{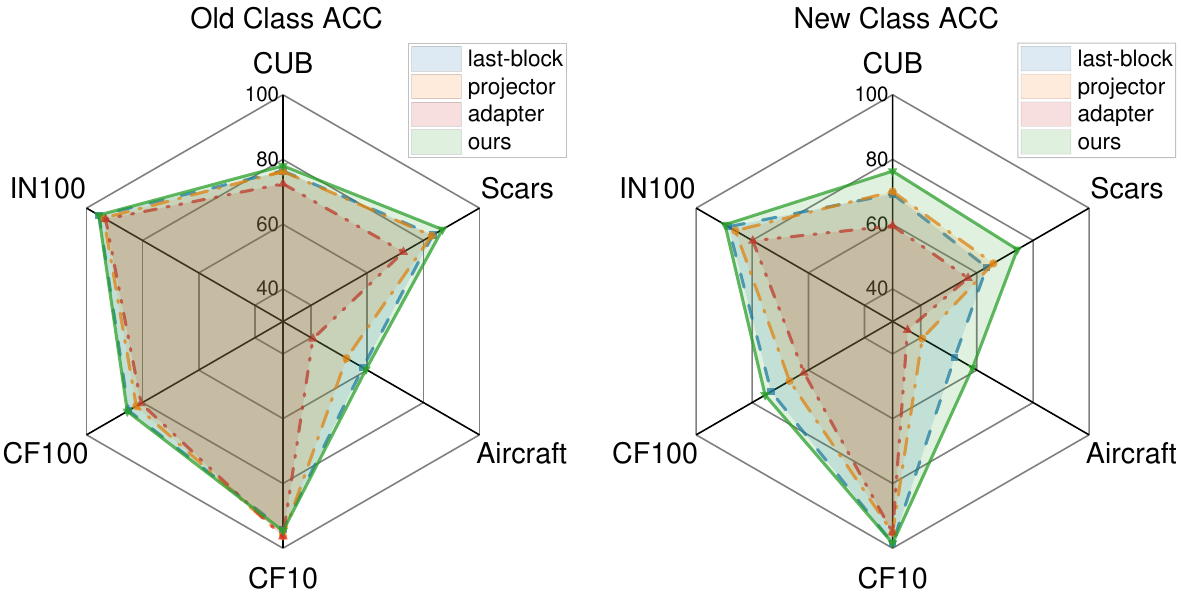}
    \caption{Different ViT finetune strategies.
    }
    \label{fig:diff_ft}
    \vspace{-6mm}
\end{figure}

\myPara{Different ViT fine-tuning strategies.}
GCD~\cite{vaze2022generalized} and SimGCD~\cite{wen2023parametric} propose to build the classifier on post-backbone features
instead of post-projector. Because the ViT backbone of CLIP contains a lot of knowledge learned from substantial image-text pairs, and the projector plays a role in modal alignment, it's essential to compare the effects of different ViT finetune strategies.
As shown in Fig.~\ref{fig:diff_ft}, we conduct multiple evaluations with last-block fine-tuning, projector fine-tuning, and adapter~\cite{gao2021clip} fine-tuning strategies. Though simply fine-tuning the projector can gain a higher accuracy across CUB and Aircraft datasets, it falls behind the last-block fine-tuning method for generic datasets. Overall, our~\OURS~perfoms the best among all methods.
For a fair comparison, we 
select the last-backbone fine-tuning strategy for baseline methods and our dual-branch multi-modal learning across all datasets except projector fine-tuning for ImageNet-1K. 

\myPara{Additional baseline results.}
As shown in \cref{subtab:p_ssb}.
We  provide results of 
PromptCAL-CLIP on 
three fine-grained datasets and three image classification generic datasets. 
For three fine-grained datasets, our method outperforms PromptCAL-CLIP on all datasets and classes. In particular, 
we surpass PromptCAL-CLIP 
by 11.5\%, 4.5\%, and 4.4\% on `All' classes of CUB, Stanford 
Cars, and Aircraft, respectively.  As for the generic datasets, our method
surpasses PromptCAL-CLIP on all datasets and achieves the best results on CIFAR-100 and ImageNet-100 datasets.

\myPara{Error bars for main results.}
The experimental results presented in the paper are the averages of three independent repeated runs.  We provide the performance standard deviation of our main results on all evaluation
datasets with three runs in \cref{tab:std}.

\begin{table}[b]
\centering
  \vspace{-2mm}
  \setlength{\tabcolsep}{9pt} 
\resizebox{0.45\textwidth}{!}{
\begin{tabular}{lccc}
\toprule
Dataset & All & Old & New \\
\midrule
CIFAR10        & 97.2$\pm$0.1 & 94.6$\pm$0.1  & 98.5$\pm$0.1   \\
CIFAR100       & 82.1$\pm$0.4   &  85.5$\pm$0.5   &  75.5$\pm$0.5   \\
ImageNet-100       &  91.7$\pm$0.3   & 95.7$\pm$0.0   &89.7$\pm$0.4   \\
CUB       & 77.0$\pm$0.5   &  78.1$\pm$1.6   & 76.4$\pm$1.2   \\
Stanford Cars       & 78.5$\pm$1.3   &  86.8$\pm$1.5   & 74.5$\pm$2.2   \\
FGVC-Aircraft       & 58.9$\pm$1.2   & 59.6$\pm$0.6   & 58.5$\pm$1.8  \\
Herbarium 19    & 49.7$\pm$0.4   & 64.5$\pm$0.8   & 41.7$\pm$0.8  \\
ImageNet-1K       & 62.4$\pm$0.0   & 74.0$\pm$0.2   &  56.6$\pm$0.1  \\
ImageNet-R       & 58.1$\pm$2.4   & 78.8$\pm$0.5   &  47.0$\pm$3.9  \\
\bottomrule
\end{tabular}}
\vspace{-2mm}

\caption{The standard deviation of our method.}
\label{tab:std}
\end{table}

\myPara{Results of two branches.}
We report the results of visual and text branches for `All' classes across six datasets in \cref{tab:2branches}. For 2 generic datasets (CIFAR10 and ImageNet-100), 
though the text branch does not achieve state-of-the-art performance, it still exhibits great performance. For 2 fine-grained datasets
(CUB and Stanford Cars), both visual and text branches 
 outperform previous methods by a large margin, while the visual branch performs better.  For 2 challenging datasets
(ImageNet-1K and ImageNet-R), both visual and text branches achieve remarkable results. 
Due to the challenging datasets comprising a significant number of unknown classes (ImageNet-1k dataset) or diverse visual concepts within the same class (ImageNet-R dataset), the consistency in text information for the same class contributes to the potentially higher discriminative power of the text branch, leading to better performance. 
\begin{table}[ht]

\vspace{-2mm}

\centering
\setlength{\tabcolsep}{9pt} 
\resizebox{.45\textwidth}{!}{
\begin{tabular}{lcc}
\toprule
Dataset & Visual Branch & Text Branch  \\
\midrule
CIFAR10        & 97.2$\pm$0.1 &   95.1$\pm$0.0   \\
ImageNet-100       &  91.7$\pm$0.3   &   90.1$\pm$0.1   \\
CUB       & 77.0$\pm$0.5   &   73.6$\pm$0.8    \\
Stanford Cars       & 78.5$\pm$1.3   &   73.1$\pm$0.6   \\
ImageNet-1K       & 62.4$\pm$0.0   &  63.5$\pm$0.1    \\
ImageNet-R       & 58.1$\pm$2.4   &   62.6$\pm$0.9    \\
\bottomrule
\end{tabular}}
\vspace{-2mm}
\caption{The results of two branches.}
\vspace{-4mm}
\label{tab:2branches}
\end{table}  

\begin{figure}[ht]
 \vspace{-1mm}
	\centering
		\includegraphics[width=0.45\linewidth]{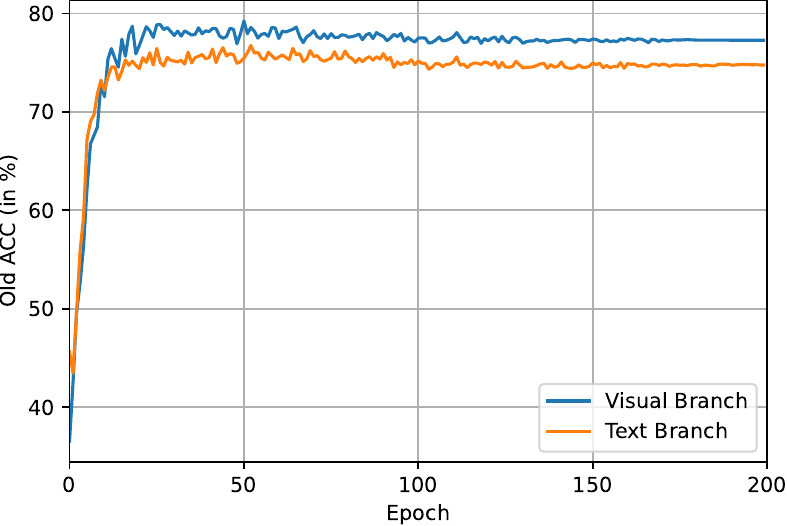}
        \hspace{5mm}
	\includegraphics[width=0.45\linewidth]{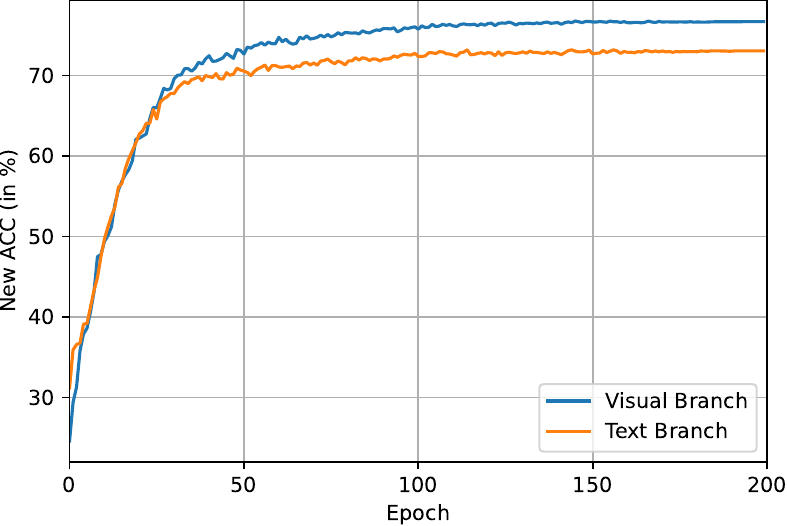}
 \vspace{-2mm}
 
        \caption{Performance evolution of two branches throughout the model learning process on the CUB dataset.}
        \label{fig:cub_2branches}
 \vspace{-4.5mm}
\end{figure}

We also provide the performance evolution of two branches throughout the model learning process on the CUB dataset (see in ~\cref{fig:cub_2branches}), the mutual promotion and fusion of the two branches resulted in excellent outcomes. In our experiments, we consistently and simply select the results from the visual branch.

\begin{table}[!b]
  \centering

  \vspace{-4.5mm}
\setlength{\tabcolsep}{2pt} 
  \resizebox{\columnwidth}{!}{
  \begin{tabular}{lccccc}
    \toprule
     Method & CIFAR10 & CIFAR100 & ImageNet-100 & CUB & Stanford Cars  \\
    \midrule
    Ground truth & 10 & 100 & 100 & 200 & 196  \\
     GCD-CLIP & 5~(50\%) & 94~(6\%) & 116~(16\%) & 212~(12\%) & 234~(19\%)  \\
     $+$TES  & 8~(20\%) & 97~(3\%) & 109~(9\%) & 212~(12\%) & 220~(12\%) \\
    \bottomrule
  \end{tabular}
  }
  \vspace{-2mm}
        \caption{Estimation of class number in unlabelled data. The table shows the estimated number and the error.} 
  \label{tab:estimation}
\end{table}

\myPara{Results with the estimated number of classes.}
Vaze \etal~\cite{vaze2022generalized} provides an off-the-shelf
method to estimate the class number in unlabelled data. We 
 introduce text embeddings generated by our TES into the off-the-shelf method by simply
concatenating text and image features before class number Estimation.
As shown in \cref{tab:estimation}, multi-modal features can estimate a more accurate class, demonstrating our multi-modal method is effective in category number estimation.
Following previous works, we assume the number of classes for each dataset is known and provide experimental results in the main paper.~\cref{tab:est_c} shows the results using the estimated number of classes for CUB and SCars datasets.

\begin{table}[t]
\small

    \centering
     \resizebox{\columnwidth}{!}{
\begin{tabular}{lccccccc}
\toprule
\multirow{2}[3]{*}{Method}                                   & \multirow{2}[3]{*}{Known $C$} &\multicolumn{3}{c}{CUB} &\multicolumn{3}{c}{Stanford Cars} \\
 \cmidrule(lr){3-5} \cmidrule(lr){6-8}
& & All  & Old  & New  & All  & Old  & New \\
\midrule
\textbf{GET}            & \ding{51}          & {77.0} & {78.1} & {76.4} & {78.5} & {86.8} & {74.5}\\
\textbf{GET}            & \ding{55}\,(w/ Est.)   & 75.6 &75.9&75.5&76.8&87.6&71.6 \\
\bottomrule
\end{tabular}}
  \vspace{-2mm}
\caption{Results with the estimated number of classes}
  \vspace{-6mm}
\label{tab:est_c}
\end{table}

\myPara{Computation complexity analysis.}
\cref{tab:cca} shows the computation complexity. Our TES uses a frozen visual encoder and stage 2  finetunes the last block in another visual encoder, thus the 2 stages share the same visual encoder for the first 11 blocks but a different last block, resulting in a low computational complexity increase.

\begin{table}[ht]
\centering

\vspace{-3mm}

\setlength{\tabcolsep}{2pt} 
\resizebox{.45\textwidth}{!}{
\begin{tabular}{lccc}
\toprule
&   \multicolumn{1}{c}{Inference Time} &   \multicolumn{1}{c}{Learnable Params} &   \multicolumn{1}{c}{FLOPs} \\
Methods &   (s/per img)  &  (M)  &   (G) \\
\midrule
SimGCD-CLIP        & $\text{5.2} \times \text{10}^{-\text{3}}$  & 13.4 & 35.2  \\
\OURS (ours)      & $\text{5.2} \times \text{10}^{-\text{3}}$   &  15.6    &  38.6   \\
\bottomrule
\end{tabular}}
\vspace{-3mm}
\caption{Computation complexity analysis.}
\label{tab:cca}
\vspace{-4mm}
\end{table}

\myPara{The anchor prototypes.}
In the CICO, the anchor prototypes are calculated by averaging the features of labeled anchor samples, making them more dynamic compared to directly using the prototype classifier $\eta$. The ablation on CUB is shown in~\cref{tab:anchor}.

\begin{table}[h]

\vspace{-3mm}
\centering
\setlength{\tabcolsep}{10pt}
\resizebox{\columnwidth}{!}{
\begin{tabular}{lccc}
\toprule
Methods & All & Old & New \\
\midrule
use classifier $\eta$      &  76.3&77.6&75.6   \\
use anchors (ours)      & 77.0&78.1&76.4   \\
\bottomrule
\end{tabular}
}
%
\vspace{-3mm}
\caption{The anchor prototypes.}
\vspace{-5mm}
\label{tab:anchor}
\end{table}

\begin{table}[!b]
\vspace{-6mm}

\centering
\setlength{\tabcolsep}{9pt}
\resizebox{\columnwidth}{!}{

\begin{tabular}{lccc}
\toprule
Methods & All & Old & New \\
\midrule
SimGCD-CLIP        & 83.1$\pm$7.4 & 99.2$\pm$0.3  & 75.1$\pm$10.9   \\
\OURS (ours)      & 90.0$\pm$1.9   &  99.2$\pm$0.2   &  85.5$\pm$2.8   \\
\bottomrule
\end{tabular}
}
%
\vspace{-3mm}
\caption{The results on Clevr-4 (Texture) in 5 runs.}
\label{tab:clevr4}
\end{table}

\myPara{Experiments on the Clevr-4 dataset}
Recently,~\cite{vaze2023clevr4} presented  a synthetic
dataset Clevr-4  to examine whether the GCD method can extrapolate the taxonomy specified by the labeled set. Most attributes of Clevr-4, such as shape, color, and count, are easily clustered (achieving close to 99\% accuracy with CLIP). However, texture attributes pose a certain level of challenge. Therefore, we evaluate our method  on the texture attributes of Clevr-4.  
As shown in \cref{tab:clevr4}, 
 our method achieves higher accuracy and lower standard deviation compared to SimGCD-CLIP, proving that the GCD method can cluster data at specified levels based on the constraint of labeled text information, which is worthy of attention and exploration.

\begin{table*}[!t]
\centering
\setlength{\tabcolsep}{3pt}
\resizebox{\textwidth}{!}{%
\begin{tabular}{lcccccccccccccccccc}
\toprule
&   \multicolumn{3}{c}{CUB} & \multicolumn{3}{c}{Stanford Cars} & \multicolumn{3}{c}{FGVC-Aircraft} &    \multicolumn{3}{c}{CIFAR10} & \multicolumn{3}{c}{CIFAR100} & \multicolumn{3}{c}{ImageNet-100}
\\
\cmidrule(rl){2-4}\cmidrule(rl){5-7}\cmidrule(rl){8-10}\cmidrule(rl){11-13}\cmidrule(rl){14-16}\cmidrule(rl){17-19}
Method        & All  & Old  & New  & All  & Old  & New  & All  & Old  & New & All  & Old  & New  & All  & Old  & New  & All  & Old  & New  \\
\midrule
WordNet~\cite{wordnet}  &41.8 &35.2 &45.1 & 26.5 & 21.9 & 29.0 & 16.5 & 13.3 & 18.2  & 18.0 & 18.6 &  17.8 & 18.6& 18.9& 18.8& 28.8& 39.1 & 23.6 
 \\
CC3M~\cite{sharma2018conceptual}  & 20.8  & 20.8 & 20.9& 18.7& 19.0 & 18.5 & 15.4& 11.6 &17.3 &  8.1 & 8.3 & 8.1 & 13.0 & 13.6 & 11.6 &8.9 &12.1 & 7.2\\

\bottomrule
\end{tabular}}
\vspace{-2mm}
\caption{Results (\%) of retrieval-based approach.} 
\label{subtab:retri}
\vspace{-3mm}
\end{table*}

\myPara{Retrieval baselines.}
To address the challenge of missing class names, another method might involve utilizing a knowledge base of potential class names (nouns) and then using CLIP to retrieve names from this corpus. Images that share the same retrieved name could be grouped together, and clustering accuracy could then be measured based on these groupings. Therefore, this retrieval-based approach serves as an important baseline.~\cref{subtab:retri} shows the results of retrieval-based approach, using WordNet~\cite{wordnet} and CC3M~\cite{sharma2018conceptual} as corpus.

\myPara{The impact of hyper-parameter $\lambda_c$.} In our method, we set $\lambda_c$
 to 1 for all datasets to prevent over-tuning.~\cref{tab:lam_c} shows the ablation of the impact of $\lambda_c$.

\begin{table}[h]
\small
\vspace{-2mm}

    \centering
\setlength{\tabcolsep}{6pt}      
     \resizebox{\columnwidth}{!}{
\begin{tabular}{lcccccc}
\toprule
\multirow{2}[3]{*}{$\lambda_c$}                                   &\multicolumn{3}{c}{CUB} &\multicolumn{3}{c}{Stanford Cars} \\
 \cmidrule(lr){2-4} \cmidrule(lr){5-7}
&  All  & Old  & New  & All  & Old  & New \\
\midrule
0.5                    & 76.3 & 74.7 & 77.1 & 79.0 & 88.7 & 74.3\\
1                    & {77.0} & {78.1} & {76.4} & {78.5} & {86.8} & {74.5}\\
1.5          &  75.3 &75.9 & 75.0 &  79.6  & 90.7  &  74.2\\
2         &  75.0 & 77.3&73.8 &79.0 & 86.8 & 75.3\\
\bottomrule
\end{tabular}}
  \vspace{-2mm}
\caption{The impact of hyper-parameter $\lambda_c$}
  \vspace{-3mm}
\label{tab:lam_c}
\end{table}

\section{Pseudo-code}
\label{ps}
The training procedure of the proposed GET is presented in~\cref{algo}. 

\begin{center}
\begin{algorithm}[t]
\algsetup{linenosize=\tiny}
\DontPrintSemicolon
\small

  \KwInput{Training dataset $\mathcal{D} = \mathcal{D}_l \cup \mathcal{D}_u$, a FC layer $l(\cdot | \theta_{\text{t}})$ and a MLP layer $g(\cdot | \theta_{\text{m}})$,
  fixed CLIP's image encoder $E_i$ and text encoder $E_t$, 
  a trainable image encoder
  $f_v(\cdot | \theta_{\text{v}})$, a  prototypical classifier $\eta(\cdot | \theta_{\text{c}})$ and a linear projection $p(\cdot | \theta_{\text{p}})$.
  }
  
  \KwOutput{Predicted label $\hat{y}_i$.
  }

  \tcc{\textbf{Stage 1: TES Training}}
    \Repeat{reaching max epochs;}{
      \For{ $(\boldsymbol{x}_i, \boldsymbol{y}_i) \in \text{each batch}$}{
      
          $\boldsymbol{z}_i^v = E_i(\boldsymbol{x}_i)$ 
          \tcp{visual embedding}  \par
          $\boldsymbol{t}_i = l(\boldsymbol{z}_i^v))  $ \tcp{pseudo text tokens}  \par
          $\boldsymbol{\hat{z}}_i^t = E_t(\boldsymbol{t}_i)$ \tcp{pseudo text embedding}  \par
          $\mathcal{L}_{align}\leftarrow$ Eq. (\textcolor{red}{5}) and Eq. (\textcolor{red}{6})\par
          $\mathcal{L}_{distill\leftarrow}$ Eq. (\textcolor{red}{7})\par
          $\mathcal{L}_{TES}= \mathcal{L}_{align} + \mathcal{L}_{distill}$ \par
          
          Back-propagation and optimize $\theta_{\text{t}}$. \par 
      }
  }

  \tcc{\textbf{Stage 2: Dual-branch  training}}
  \Repeat{reaching max epochs;}{
      \For{ $(\boldsymbol{x}_i, \boldsymbol{y}_i) \in \text{each batch}$}{
      \tcc{\textbf{Visual-branch}}
          $\boldsymbol{z}_i^v = f_v(\boldsymbol{x}_i), \ \boldsymbol{h}_i^v=g(\boldsymbol{z}_i^v), \ \boldsymbol{p}_i^v = \eta(\boldsymbol{z}_i^v) $ \par
          Compute $\mathcal{L}_{ucon}^v$ and $\mathcal{L}_{scon}^v$ by 
          replacing $\boldsymbol{h}$ in  Eq. (\textcolor{red}{1}) and Eq. (\textcolor{red}{2}) with $\boldsymbol{h}^v$ 
          
          \par
          $\mathcal{L}_\text{rep}^v\leftarrow$ Eq. (\textcolor{red}{8}) \par 
          
          Compute $\mathcal{L}_\text{cls}^v$ by 
          replacing $\boldsymbol{p}$ in  Eq. (\textcolor{red}{3}) and Eq. (\textcolor{red}{4}) with $\boldsymbol{p}^v$ 
          
          $\mathcal{L}_\text{db}^v\leftarrow$Eq. (\textcolor{red}{9}) \par
          
          \tcc{\textbf{text-branch}}
          $\boldsymbol{\hat{z}}_i^{t} = E_t(l(E_i(\boldsymbol{x}_i))) $ \par
         $\boldsymbol{\hat{z}}_i^{tl} = p(\boldsymbol{\hat{z}}_i^t), \ \boldsymbol{h}_i^t=g(\boldsymbol{\hat{z}}_i^{tl}), \ \boldsymbol{p}_i^t = \eta(\boldsymbol{\hat{z}}_i^{tl}) $ \par
         $\mathcal{L}_\text{db}^t =  \mathcal{L}_\text{rep}^t + \mathcal{L}_\text{cls}^t$ \par
         Compute the multi-modal mean entropy regularization $H_{mm}$ \par
         \tcc{\textbf{CICO}}
         Calculate the visual and text anchors $\mathcal{P}_v,\mathcal{P}_t$  \par
         Compute the instance relationships by Eq. (\textcolor{red}{10}) \par
         $\mathcal{L}_\text{CICO}\leftarrow$ Eq. (\textcolor{red}{11}) \par
            $\mathcal{L}_\text{Dual}\leftarrow$ Eq. (\textcolor{red}{12})
          \par

         Back-propagation and optimize $\theta_{\text{v}}, \ \theta_{\text{m}}, \  \theta_{\text{c}}, \  \theta_{\text{p}}$.

      }
  }
  \Return{$\hat{y}_i = \eta(f_v(\boldsymbol{x}_i))$.} \par
\caption{Pseudocode for \mbox{{~\OURS}}.}
\label{algo}
\end{algorithm}
\end{center}

\section{Different pre-trained models}
\label{ssec:pm}

In this section, we perform an extensive empirical investigation to explore the impact of different types of pre-trained models on GCD clustering performance,  which clearly demonstrates that different types of backbones exhibit varying biases across different datasets, classes, and even paradigms. We choose DINO~\cite{caron2021emerging}, 
which is based on teacher-student learning; 
MoCo v3~\cite{chen2021empirical}, 
based on contrastive learning; iBOT~\cite{zhou2021ibot}, based on contrastive masked image modeling; and CLIP~\cite{radford2021learning}, which is based on vision-language contrastive learning. 

\begin{figure*}[!t]
    \centering
    \includegraphics[width=\textwidth]{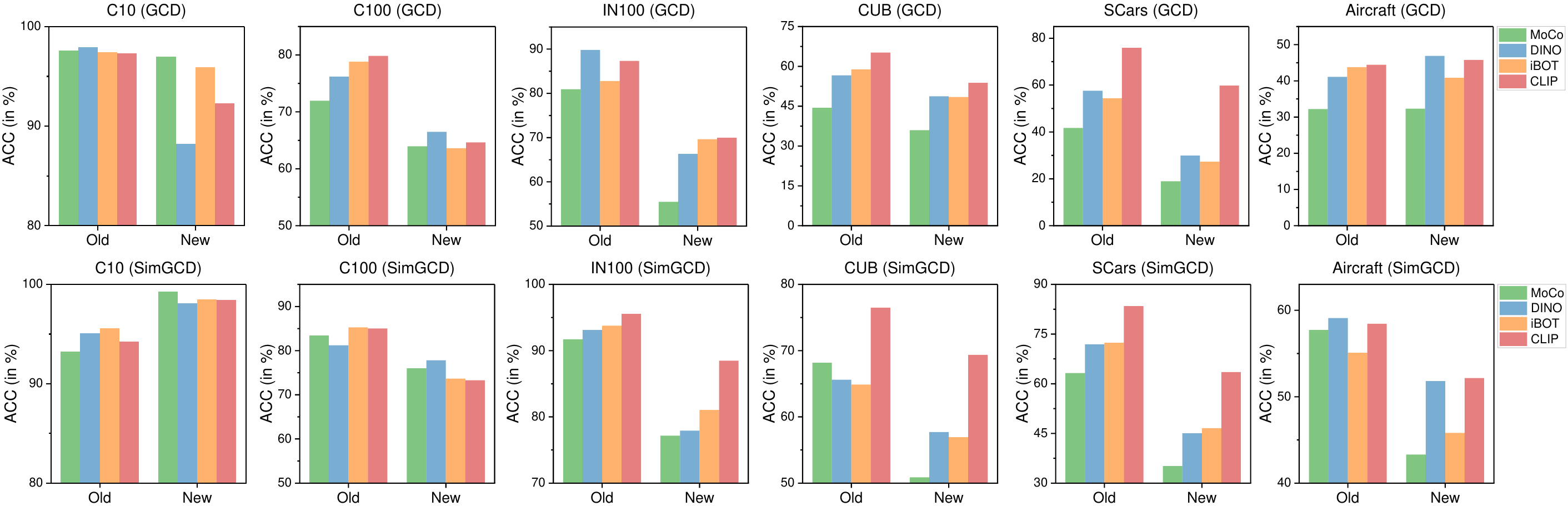}
\caption{The results of GCD and SimGCD with different backbones across six datasets. }
    \label{fig:diff_backbone}
\end{figure*}

We first evaluate the results of GCD and SimGCD across different types of pretraining models. 
As shown in Fig.~\ref{fig:diff_backbone}, 
different types of backbones exhibit varying biases across different datasets, classes, and even paradigms. For example, iBOT outperforms DINO in non-parametric GCD, but DINO excels in parametric GCD. MOCO demonstrates the strongest category discovery ability for the  CIFAR dataset.
CLIP performs exceptionally well across all datasets, yet struggles with low-resolution CIFAR data in parametric GCD.

\begin{figure*}[!t]
    \centering
    \includegraphics[width=.98\textwidth]{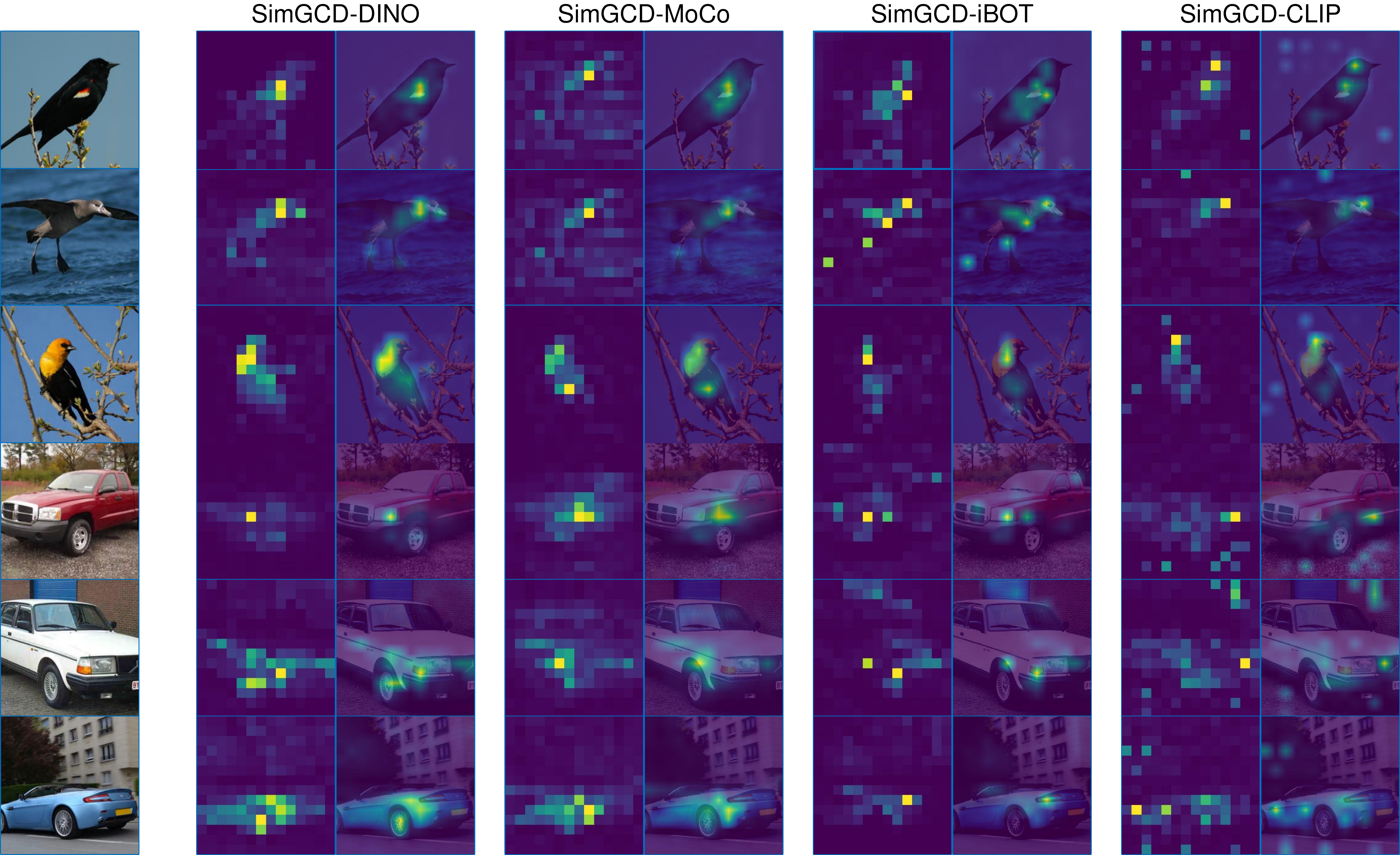}
    \caption{Attention map of class tokens on CUB (first three rows) and StanfordCars (last three rows) datasets. Each row displays the attention areas and attention maps for each image of SimGCD~\cite{wen2023parametric} with different backbone models.
    }
    \label{fig:diff_b_am}
    \vspace{-3mm}
\end{figure*}

We then visualize and compare the attention map of class tokens of different backbones in~\cref{fig:diff_b_am}. For the CUB dataset, the DINO, iBOT, and MoCO backbones tend to focus more on the feathers of the birds, while CLIP additionally emphasizes the more discriminative head area. For the StanfordCars dataset, the DINO backbone focuses on the car light and wheel; the MoCo backbone focuses on the front fenders of the car, which is less discriminative; the iBOT backbone focuses on the car light and the car window, which is more discriminative than DINO thus leading to better results; the CLIP backbone focus on both the front of the car and global information, showcasing stronger discriminative capabilities.

A key observation is that though promising results have been achieved, different backbones, even powerful CLIP, still perform inferiorly on distinguishing certain visually similar classes, such as the classes in all fine-grained datasets. 
We argue that this is due to
current methods only utilize a single visual modality of information, another modality may potentially compensate for the lack of discriminative ability.
In the meanwhile, the potential of current GCD methods heavily relies on the generalization ability of pre-trained models, prompting us to select a more robust and realistic pre-training model. 
As a large-scale model, CLIP shows strong generalization ability on downstream tasks and strong multi-modal potential due to its image-text contrastive training, 
thus we decide to introduce it into the GCD task. 
This not only unleashes the latent potential performance of existing methods but also serves as a bridge for us to leverage multi-modal information.

\begin{figure*}[t]
    \centering
    \includegraphics[width=\textwidth]{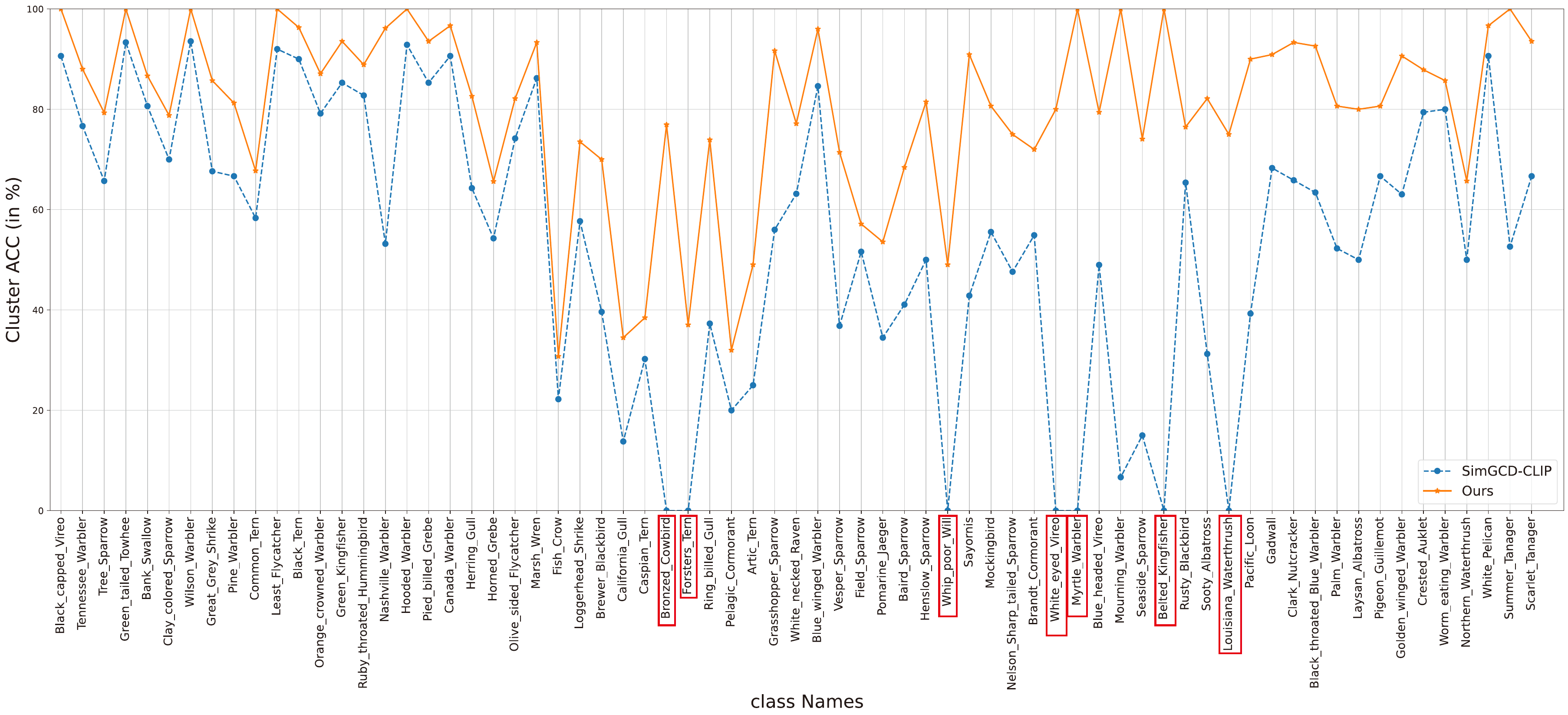}
    \caption{Cluster accuracy of SimGCD-CLIP and our \OURS~on some visually similar classes in CUB datasets. GCD methods relying solely on a single visual modality result in empty clusters(highlighted by red boxes); Our multi-modal approach \OURS~avoids empty clusters and achieves higher classification accuracy.}
    \label{fig:cub_plot}
    \vspace{-3mm}
\end{figure*}

\section{Cluster results of GET}
\label{sB}

\begin{figure*}[t]
    \centering
    \includegraphics[width=.7\textwidth]{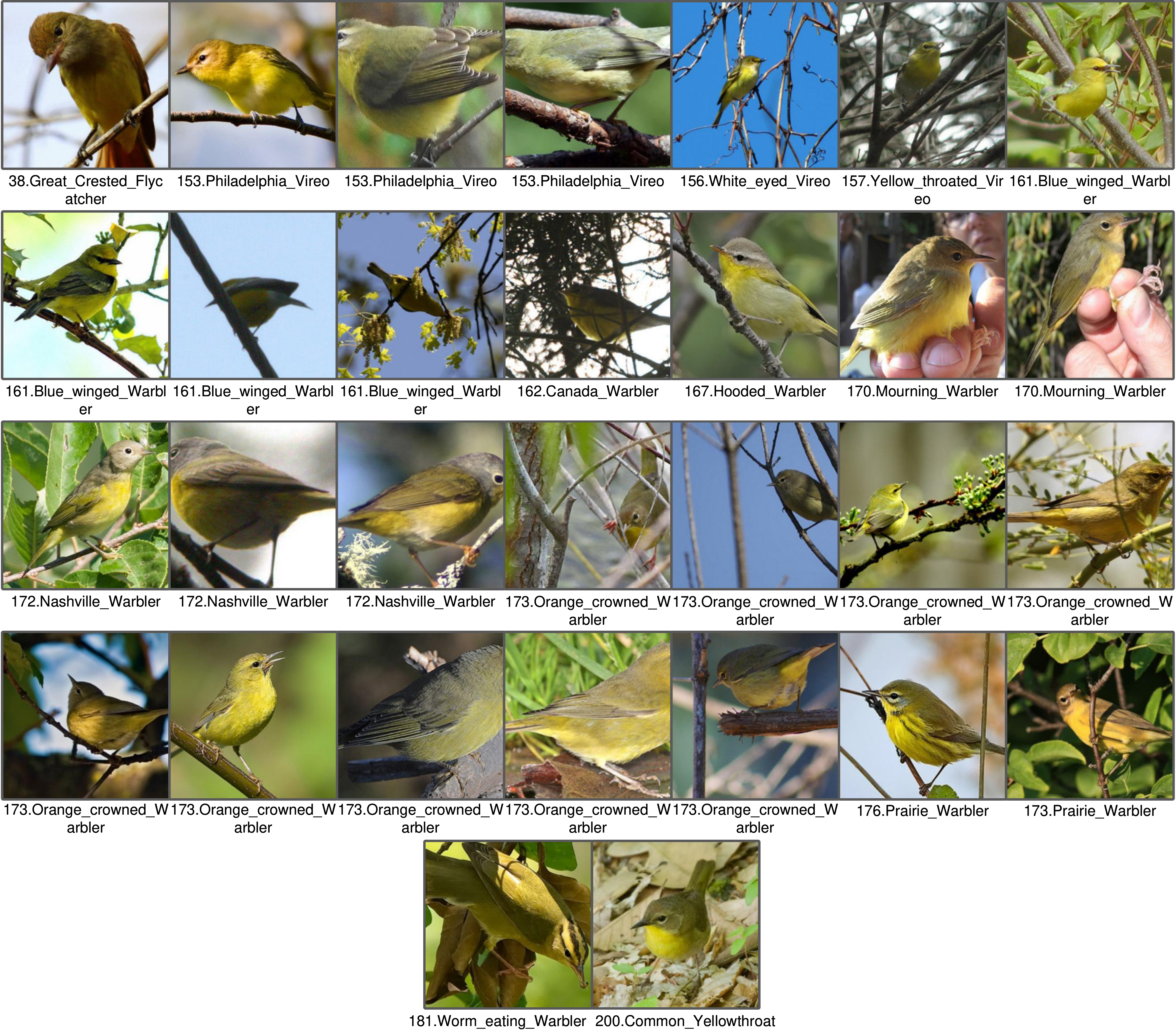}
    \caption{SimGCD-CLIP cluster results visualization for class ``Mourning Warbler'' in CUB dataset. SimGCD-CLIP categorizes birds based on shape and posture and incorrectly identifies many visually similar categories, resulting in a clustering accuracy of 6.7\% for class ``Mourning Warbler''.}
    \label{fig:cub_170_clip}
     \includegraphics[width=.7\textwidth]{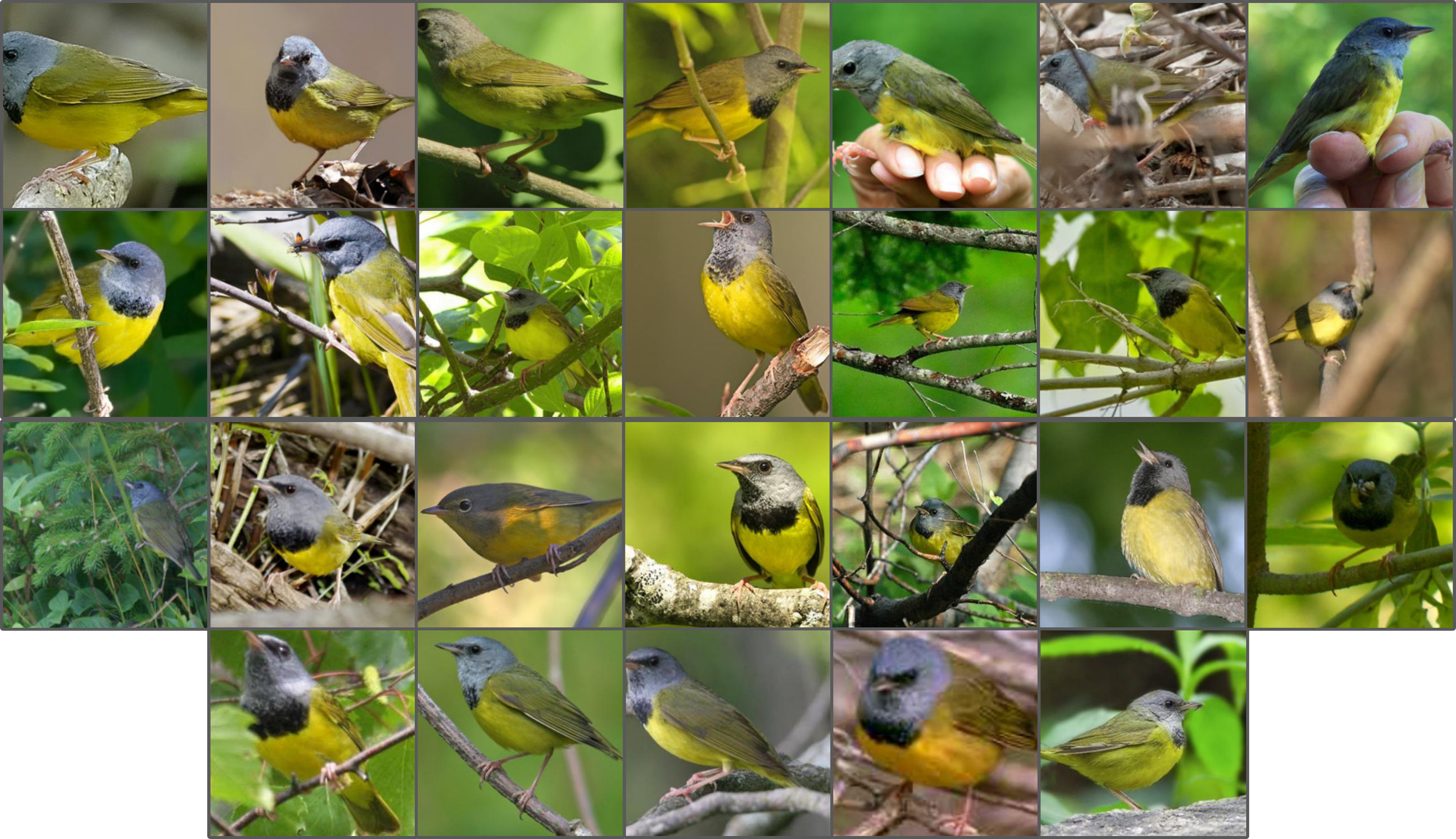}
     \caption{Cluster results visualization for class ``Mourning Warbler'' in CUB dataset of our \OURS. Our method uses multi-model information, achieving 100\% classification accuracy for this visually challenging category.
    }
    \label{fig:cub_170_ours}
\end{figure*}

As shown in \cref{fig:cub_plot}, we present the comparative cluster accuracy between our multi-modal approach and previous single-modal methods on some visually similar classes in CUB datasets. 
It is worth noting that relying solely on visual information, even with a powerful CLIP backbone, the previous method (SimGCD-CLIP)   still struggles to differentiate some categories, resulting in empty clusters.
However, leveraging the rich and discriminative text information of categories, our \OURS~achieves more accurate classification results on CUB without any empty clusters across all categories, demonstrating the importance of multi-modal information in the GCD task. 
Furthermore, we showcase the clustering results of SimGCD-CLIP (see in \cref{fig:cub_170_clip}) and our \OURS~ (see in \cref{fig:cub_170_ours}) for the 170th class, ``Mourning Warbler'',  in the CUB dataset. 
SimGCD-CLIP relies solely on visual information to categorize birds based on shape and posture,  the model categorizes many visually similar samples as ``Mourning Warbler''. Our approach, by incorporating text information, enhances the model's discriminative ability and correctly identifies all instances of the ``Mourning Warbler'' class, achieving 100\% classification accuracy for this visually challenging category.

\section{Limitations and Broader Impact}
\label{lm}
\myPara{Limitations and future works.}
A limitation of our approach is
that we treat visual and text information as equally important. In fact, some
samples may have richer and more discriminative visual information than textual
information, and vice versa. A more appropriate approach might involve enabling
the model to adaptively leverage multimodal information, autonomously assessing which modality’s information is more crucial. We will delve deeper into this
aspect in our future work.

\myPara{Broader impact.}
Our approach introduces text information into the GCD task through a novel text
embedding synthesizer module, extending the GCD to a multi-modal paradigm without extra corpus or models. We believe that the introduction of TES will encourage future research in solving GCD in a multi-modal manner.